\documentclass{IEEEtran}%

\usepackage{graphicx}%
\usepackage{subcaption}%
\usepackage{multirow}%
\usepackage{amsmath,amssymb,amsfonts}%
\usepackage{amsthm}%
\usepackage{mathrsfs}%
\usepackage[dvipsnames,table]{xcolor}%
\usepackage{textcomp}%
\usepackage{manyfoot}%
\usepackage{booktabs}%
\usepackage{algorithm}%
\usepackage{algorithmicx}%
\usepackage{algpseudocode}%
\usepackage{listings}%
\usepackage{lipsum}
\usepackage{glossaries}
\usepackage{duckuments}
\usepackage{hyperref}
\usepackage[switch]{lineno} %
\usepackage{xurl} %
\usepackage{xspace}
\usepackage{pifont}%
\usepackage{tikz}
\usetikzlibrary{positioning, arrows.meta, fit, backgrounds}

\newcommand\etal{\emph{et al.}\xspace}
\newcommand\ie{\emph{i.e.},\xspace}

\newcommand{\tech}[1]{{\small{\textsf{#1}}}}

\newcommand{\fnunit}{\footnotesize\color{gray}}
\tikzset{
    base/.style     = {draw=black!72, text width=2.0cm, align=center, minimum height=7.5mm,
                       font=\footnotesize, inner sep=2pt},
    inp/.style      = {base, fill=blue!10},
    calcnode/.style = {base, fill=orange!10},
    pwr/.style      = {base, fill=green!12},
    hwn/.style      = {base, fill=yellow!18},
    outp/.style     = {base, fill=red!8, text width=2.1cm},
    lane/.style     = {draw=gray!55, dashed, rounded corners, inner sep=2mm},
    lname/.style    = {anchor=south west, font=\itshape\footnotesize, gray, inner sep=1pt},
    flow/.style     = {-{Stealth}, thick, black!62},
    hwflow/.style   = {-{Stealth}, thick, dashed, black!45},
    outer/.style    = {draw=gray!70, rounded corners, inner sep=2.5mm},
}

\newcommand{\cmark}{\ding{51}}%

\newacronym[
    \glslongpluralkey={\emph{Large Language Models}},
    \glsshortpluralkey={LLMs}
]{llm}{LLM}{\emph{Large Language Model}}

\theoremstyle{plain}%

\theoremstyle{remark}%

\theoremstyle{definition}%

\begin{document}

\title{Measured from the feed: the unsustainable growth of AI video generation}

\author{
\IEEEauthorblockN{Tristan Coignion\IEEEauthorrefmark{1}, A.~Bugeau\IEEEauthorrefmark{2}\IEEEauthorrefmark{3}, G.~Guennebaud\IEEEauthorrefmark{2}\IEEEauthorrefmark{4}, and A.-L.~Ligozat\IEEEauthorrefmark{5}}

\IEEEauthorblockA{\IEEEauthorrefmark{1}Univ.~Lille, CNRS, Centrale Lille, UMR 9189 CRIStAL, F-59000 Lille, France}

\IEEEauthorblockA{\IEEEauthorrefmark{2}Univ.~Bordeaux, CNRS, Bordeaux INP, LaBRI, UMR~5800, Talence, France}

\IEEEauthorblockA{\IEEEauthorrefmark{3}Institut Universitaire de France (IUF)}

\IEEEauthorblockA{\IEEEauthorrefmark{4}Inria}

\IEEEauthorblockA{\IEEEauthorrefmark{5}Universit\'e Paris-Saclay, CNRS, ENSIIE, LISN, 91400, Orsay, France}

\thanks{Corresponding author: Tristan Coignion, tristan@coignion.fr}
}

\maketitle

\begin{abstract} %
Text-to-video generation runs at commercial scale, but its broader
environmental impacts remain, to our knowledge, understudied. Providers
rarely disclose video volumes or energy needs; we measured the
yearly video output of two commercial platforms and built a prospective model of such services' absolute footprint
from training to storage.
Our projections suggest that reducing emissions depends on both limiting growth in video generation and achieving substantial efficiency gains.
Under continued growth, emissions rise nearly seventy-five-fold by 2030 despite technological improvements.
When usage remains at 2026 levels, the same improvements reduce emissions by about two-fifths.
These estimates are a conservative lower bound: they cover one AI service
among many; other services and uses would push the total higher.

\end{abstract}

The rapid growth of artificial intelligence raises concerns about the environmental footprint of the infrastructure that sustains it: this footprint is increasingly documented in aggregate, but its breakdown by usage type remains invisible.
AI services run in data centres whose electricity demand is rising fast, whose cooling consumes large quantities of water, and whose dedicated hardware is itself a growing manufacturing burden: the carbon footprint of producing one graphics card has roughly quadrupled over the last decade, to about 300~kg~CO\textsubscript{2}eq~\cite{morandRisingUnsustainabilityAI2026}.
Within this growth, one modality stands out for its resource cost: video generation.
Delavande \etal found that generating a short video from text consumes roughly 90~Wh: nearly 30$\times$ the energy of image generation and 45,000$\times$ that of text classification~\cite{delavandeVideoKilledEnergy2025}.

Video generation services do not run on the user's device: each request runs on racks of AI accelerators inside dedicated data centres.
The resulting pressure is localised: nearly two-thirds of the new US data centres built since 2022 lie in high water-stress regions~\cite{privetteDataCentersWater2026}, where water withdrawals and thermal discharges can threaten freshwater biodiversity~\cite{jagerDataCentersThreat2026}.
As AI video generation grows~\cite{kuaishouKlingARR240M2026}, such facilities may put an ever bigger strain on local energy grids and water resources.

Once experimental, AI video generation now serves entertainment~\cite{sunAutomaticSocialMedia2026, wangBasedRunwayGen42025}, personalized targeted advertisements~\cite{kapoorFrontiersGenerativeAI2025}, education videos~\cite{hai2025rapid}, as well as mass propaganda and misinformation~\cite{kumarPeepingFutureUnderstanding2025, samsonComputationalPropagandaMisinformation2025}.
In early 2026, the leading platform Kling AI surpassed 12 million monthly active users~\cite{dailyKlingSurpasses122026}.
Professional production is following: Netflix reported in July 2026 that roughly 300 of its titles use generative-AI workflows~\cite{netflixGenAIWorkflows2026}, and Runway reports enterprise adoption across hundreds of companies~\cite{runwayAIMediaReport2026}.

While current studies estimate the energy consumption and carbon footprint of generating individual videos~\cite{jeghamLightsCameraCarbon2026,delavandeVideoKilledEnergy2025,liCarbonMotionCharacterizing2024}, these per-unit figures have limits.
First, they are intensity indicators: video streaming shows that
increasing the resolution increases the total energy consumption while lowering the energy per gigabyte.
The efficiency indicator can thus improve while the service's absolute impact increases~\cite{guennebaudEnergyConsumptionDataTransfer2024}.
Second, they describe a single generation, not a deployed service: they say nothing about the number of videos a service produces, nor about the infrastructure that number requires.
Third, they cover few impact categories: a per-video energy figure says nothing about the cooling water or the hardware production these services require.
The broader environmental impacts of AI video generation as a technology, \ie its absolute footprint at worldwide scale and its evolution over time, are, to our knowledge, still understudied.

This study pursues two goals: estimating the direct environmental impacts of AI video generation as a deployed technology, at worldwide scale and for the present day, and exploring how these impacts could evolve to 2030 under contrasted scenarios of demand and video quality.

To make the problem tractable, we limit the scope to specialized text-to-video (T2V) generation services: platforms that produce a complete video clip from a text prompt.
We therefore exclude other applications of generative AI to video, such as editing or enhancing existing footage~\cite{runwayAleph2025,adobeGenerativeExtend2026}, avatar-based talking-head videos~\cite{synthesiaAvatars2026}, and video generation embedded in a broader creative suite~\cite{canvaMagicMedia2026}.
Impacts beyond the AI provider, such as video streaming, are also excluded.
Since the effects we exclude would only add to the computing-related
impacts of this service, our estimates form a lower bound for those impacts.

We build a bottom-up model
of the serving infrastructure: starting from the energy and latency of a single generation, we scale these to the yearly volume of generated videos to size the required hardware, and from that hardware we derive the operational electricity use and the environmental impacts.
We anchor this volume in a measurement campaign: we crawled Sora's public feed for 54 days and combined it with HeyGen's public activity counter, which gives a measured floor for the total number of videos generated per year.
To account for uncertainties, we quantify the spread of our estimates with a Monte Carlo analysis, and report which assumptions drive the results.

Here we provide, to our knowledge, the first estimate of the worldwide direct footprint of AI video generation as a deployed technology.
We combine our two-platform measurements with the generation counts disclosed by six providers into a 2026 volume of 6.2 to 9.5 billion single-output videos per year.
The model reports four indicators together: operational electricity, greenhouse gas emissions, operational water, and embodied abiotic resource depletion.
In our median 2026 estimate, video generation consumes 724~GWh of electricity and emits 124.3~kt~CO\textsubscript{2}eq, a footprint dominated by inference rather than by training.
Under our \emph{Reference scenario}, emissions grow nearly seventy-five-fold by 2030; freezing demand and video quality would cut them by about two-fifths.
Holding emissions flat would instead require improving the inference stack by about 75\% per year; the fastest rate we test, 30\% per year, leaves 2030 emissions more than twenty-five times above their 2026 level, so efficiency gains alone cannot offset demand growth.

\section{Results}\label{sec:results}

\subsection{How many AI videos are generated per year?} %

To size the annual volume of generated videos, we combine our own
measurements on two platforms with the figures disclosed by some of the largest
providers (Section~\ref{subsec:volumetry}).
Our own measurements cover two services.
Sora, OpenAI's text-to-video model, maintained a public feed that listed every public video with its resolution, duration and prompt; HeyGen, whose text-to-video service displayed a real-time counter of generated videos on its homepage.

We crawled Sora's public feed for 54 days between February and April 2026, recording
3.3M public videos, which annualizes to 22M per year; OpenAI
discontinued Sora's web and app experiences shortly after the crawl
ended~\cite{openaiSoraDiscontinuation2026}.
Likewise, from a 90-day monitoring of HeyGen's counter, we deduced a rate of 111M
videos per year, including private ones. Together they give a measured floor of roughly 133M videos
per year.
To our knowledge, six other providers publish generation counts, whose annualised rates sum to an observable total of about 6.2 billion single-output videos per year: Kling, about 360M per year~\cite{kuaishouKlingARR240M2026}; Veo in Flow, about 660M~\cite{gallegosIntroducingVeo312025}; MiniMax, about 430M; Vidu, about 267M; Higgsfield, about 560M; and Grok Imagine, about 3.8B 
at single-output equivalents, which alone accounts for most of the total volumetry in our study.
Both lines of evidence are partial, so our model draws its 2026 volume from a one-sided
triangular distribution spanning 6.2--9.5 billion videos, with its low end and mode anchored on 6.2B.

\begin{figure}[t]
    \centering
    \includegraphics[width=1\linewidth]{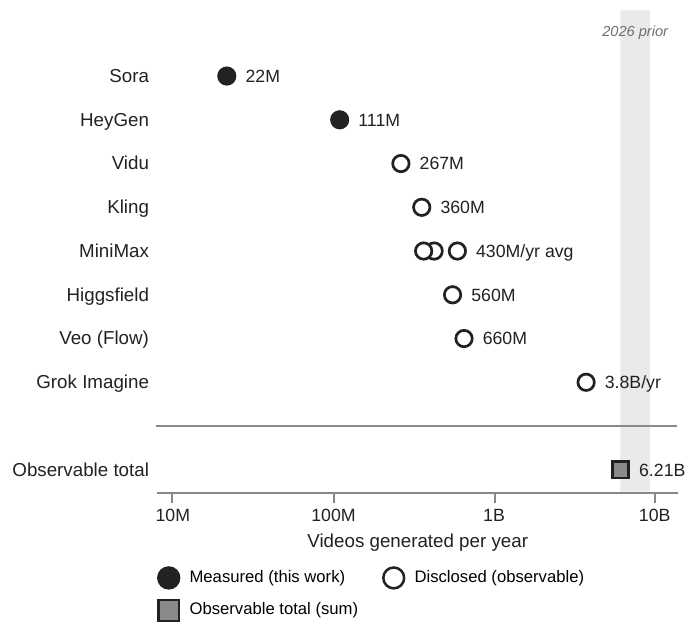}
    \caption{Annual volume of AI-generated videos. The shaded band is the
        model's 2026 prior on video volume (one-sided triangular, 6.2--9.5 billion,
        low end and mode 6.2 billion). Provider rows are grouped in two tiers:
        our measurements, and the disclosed observable rates.
    }
    \label{fig:volumetry}
\end{figure}

\subsection{Brief overview of our model}

\textbf{Service under study.}
We study one year of operation of an AI video-generation
\emph{service} and estimate the infrastructure and energy it requires.
The model covers three accounting components (\textbf{training}, \textbf{inference}, and
\textbf{storage}) and excludes the diffusion of videos to viewers
(content delivery, playback, social-media storage).
The precise boundary, including the audio exclusion and the operational/embodied split, is given in the Methods
(Section~\ref{subsec:scope}).
The model runs over the years 2026--2030 and reports per year the required GPU
fleet, the stored data, operational electricity use (consumption and peak power), and
three environmental impacts:
greenhouse-gas emissions, operational water use (facility water input per unit of IT energy,
from a data-centre WUE covering direct potable and non-potable water and excluding
indirect water use), and embodied abiotic depletion potential (ADP, kg Sb-eq).

Within this service, we represent all providers' output with one reference
generation model, treated as a black box.
One \textbf{inference} is then one video produced end-to-end by one GPU, with
durations and resolutions representative of the service's output
(Supplementary Methods~\ref{supp:volumetry}).

\textbf{Training} covers the sector's yearly retraining runs: the 20 to 40 runs per year represent every laboratory that trains AI video-generation models, at one to two trainings per laboratory (Supplementary Methods~\ref{supp:retraining}), plus a development overhead for preliminary and failed runs.

\textbf{Storage} covers the videos that providers keep on their drives over a
multi-year retention window, in replicated copies.
The Methods give the full boundary and the equipment model
(Sections~\ref{subsec:scope} and~\ref{subsec:variables}).

\textbf{Uncertainty.}
Every numeric input can carry an uncertainty distribution; we run 10{,}000
Monte Carlo simulations and report medians with p5--p95 ranges, and a
one-at-a-time sensitivity analysis ranks the inputs by influence.

\subsection{Impacts of AI video generation in 2026}

\begin{figure}[ht]
    \centering
    \includegraphics[width=1\linewidth]{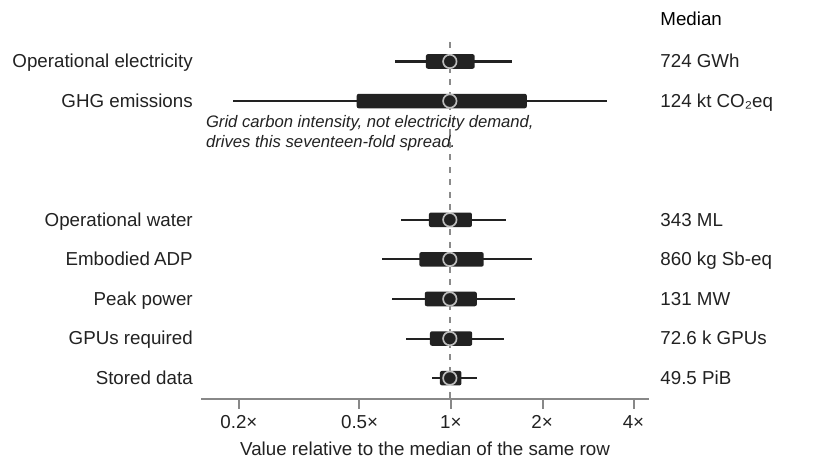}
    \caption{Estimated 2026 footprint of specialized text-to-video generation,
        with sensitivity. Each row is normalised to its own median so that
        spreads can be compared; absolute medians are printed at the right.
        Dot: median. Thick bar: p25--p75. Thin bar:
        p5--p95, over 10{,}000 Monte Carlo samples. 
    }
    \label{fig:impacts-2026}
\end{figure}

\textbf{Impact analysis.}
\autoref{fig:impacts-2026} reports our estimates for 2026, the first simulated
year, together with their uncertainty.
The 7.16 billion videos estimated for 2026 (the median of the Monte Carlo runs) require a fleet of
72.6k GPUs, consume 724~GWh of electricity, and emit 124.3~kt
CO\textsubscript{2}eq.
Under an equal per-capita distribution of the planetary
climate boundary~\cite{salaEnvironmentalSustainability2020}, these emissions
correspond to the targeted annual emissions of about 126,000 people.

Dividing total electricity by the number of videos gives an \emph{all-in} cost of
about 100~Wh per video, \ie 4.1$\times$ the 24.4~Wh of GPU energy alone that the model
charges to a single generation (a fleet-weighted average over the 2026
resolution and duration mix).
Idle fleet time, facility overheads, training and storage multiply the single-inference figure more than fourfold.

\begin{figure}[ht]
    \centering
    \includegraphics[width=0.7\linewidth]{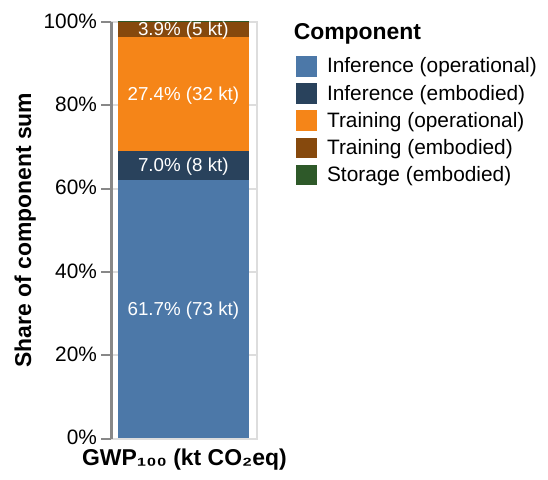}
 
    \caption{Operational and embodied components of the global-warming potential (GWP$_{100}$, kg CO\textsubscript{2}eq) of specialized text-to-video generation in 2026. Medians over 10{,}000 Monte Carlo samples; shares are of the summed component medians, which under-state the median total (124.3~kt CO\textsubscript{2}eq) by about 5\% because medians are not additive.
    The storage share (0.01\%) is too thin to display.
    }
    \label{fig:impacts-proportions-2026}
\end{figure}

\textbf{Distribution of the impacts.}
\autoref{fig:impacts-proportions-2026} breaks down global-warming potential by model component and by operational versus embodied origins; operational electricity and embodied ADP, each confined to a single scope (use-phase electricity, hardware manufacture), are decomposed in the text instead.
One observation stands out: \emph{inference dominates every indicator in 2026}.
Inference is the largest single component of the footprint: 68.7\% of the summed GHG
component medians, 66.5\% of operational electricity, and 64.0\% of embodied ADP.
Training is the second-largest component: 31.3\% of the summed GHG
component medians (27.4\% operational, 3.9\% embodied), 33.4\% of operational
electricity, and 36.0\% of embodied ADP. Storage is negligible
(under 0.2\% of every indicator).
The weight of training reflects its representation: each retraining run carries an
energy cost independent of the yearly volume, and the model draws 20--40 retraining
runs per year.
The 2026 balance is robust to the volume prior: inference dominates even at the
6.2-billion-video observable total that anchors the prior's low end, and halving the
retraining count lowers 2026 GHG emissions by 16.6\% without changing the ranking
(Supplementary Results~\ref{supp:training-constancy}).
Inference and training are still calibrated asymmetrically: inference is
anchored on a production-scale commercial model, whereas training rests on a
few published runs of small open-source T2V-models scaled by an R\&D
overhead. These shares depend on the assumptions and evidence currently available
for training and inference. The inference-dominated balance agrees with a service-level LCA of
Stable Diffusion, which likewise finds global-warming
and primary-energy impacts dominated by inference
cost~\cite{berthelotEstimatingEnvironmentalImpact2024}.

\begin{figure*}[ht]
    \centering
    \includegraphics[width=1\linewidth]{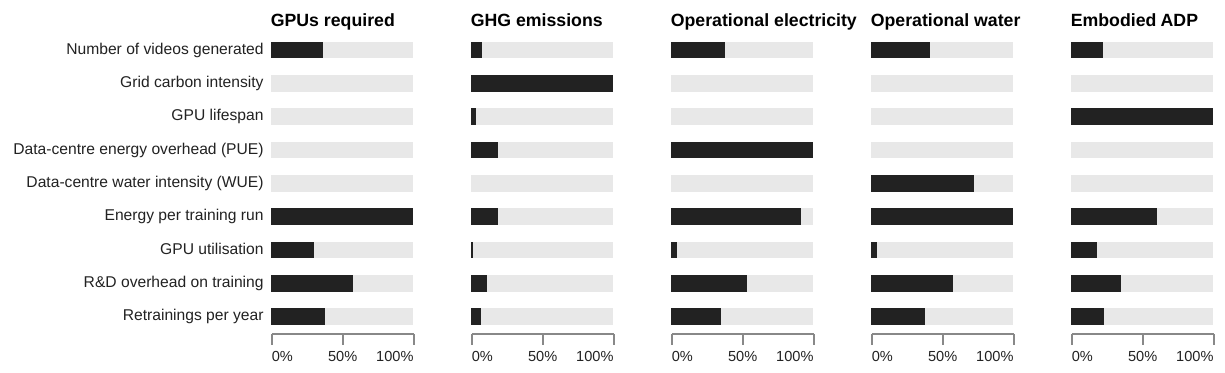}

    \caption{One-at-a-time sensitivity analysis of the nine main input parameters for the reference year 2026. Each
        input is varied alone across its sensitivity distribution while all
        other inputs stay at their medians; bar length is the resulting p5--p95
        output spread, normalised within each column to its largest value. Bars
        are comparable within a column, not across columns. This ranks the
        contribution of each assumption to output uncertainty.}
    \label{fig:tornado-2026}
\end{figure*}

\textbf{Sensitivity analysis.} 
We vary each input alone across its sensitivity distribution with all other inputs at their medians.
In \autoref{fig:tornado-2026}, the energy per training run is
the largest single contributor in two of the five columns, required GPUs and water, with
the R\&D overhead and the number of retrainings close behind. The facility parameters
lead where they act directly: PUE produces the largest spread in electricity, and the
water-use-efficiency coefficient the second-largest in water.
Greenhouse-gas emissions are the exception:
there, grid carbon intensity produces 12$\times$ the spread of video volume, so
\emph{where} the electricity comes from matters more than \emph{how much}
electricity is used.
ADP is the one column where a hardware parameter leads: GPU lifespan produces 4.4$\times$ the spread of video volume, so hardware renewal cadence appears as a strong lever to reduce ADP. %

Two more patterns emerge: the conversion coefficients keep their expected columns (carbon intensity first in GHG emissions; WUE (the EU-hyperscaler average of direct potable and non-potable water use that our WUE prior is centred on~\cite{euDataCentresEnergyPerformance2025}) second in water, behind the energy per training run; and PUE first in electricity); and the three training-related parameters appear in all five columns and lead two of them, consistent with training accounting for about a third of the 2026 footprint.

The same asymmetry appears in the 2026 distributions (\autoref{fig:impacts-2026}): electricity spans a factor of 2.4 between the 5th and 95th percentiles, against 17.1 for GHG emissions.

\subsection{Future impacts of AI video generation}\label{sec:future}

We contrast three scenarios that define what may change by 2030; their full
year-by-year parameters are given in the Methods
(Section~\ref{subsec:uq}). All scenarios share the same technology
trajectory, modelling the efficiency improvements of a real inference
stack: an inference-optimization factor falls by a cumulative $\sim$35\% over 2026--2030, about 10\% per year compounded. Because this inference optimization factor evolution is arbitrary, we made an inversion analysis below and a sensitivty analysis in the Supplementary Results~\ref{supp:opt-sensitivity}.

GPU utilisation, the fraction of time the hardware spends generating rather than
idle, rises by a cumulative $\sim$18\%; and the GPU fleet renews from 100\%
H100 in 2026 towards 95\% B200 by 2030, a newer and more energy-efficient
chip. 

\begin{figure}[t]
    \centering
    \includegraphics[width=1\linewidth]{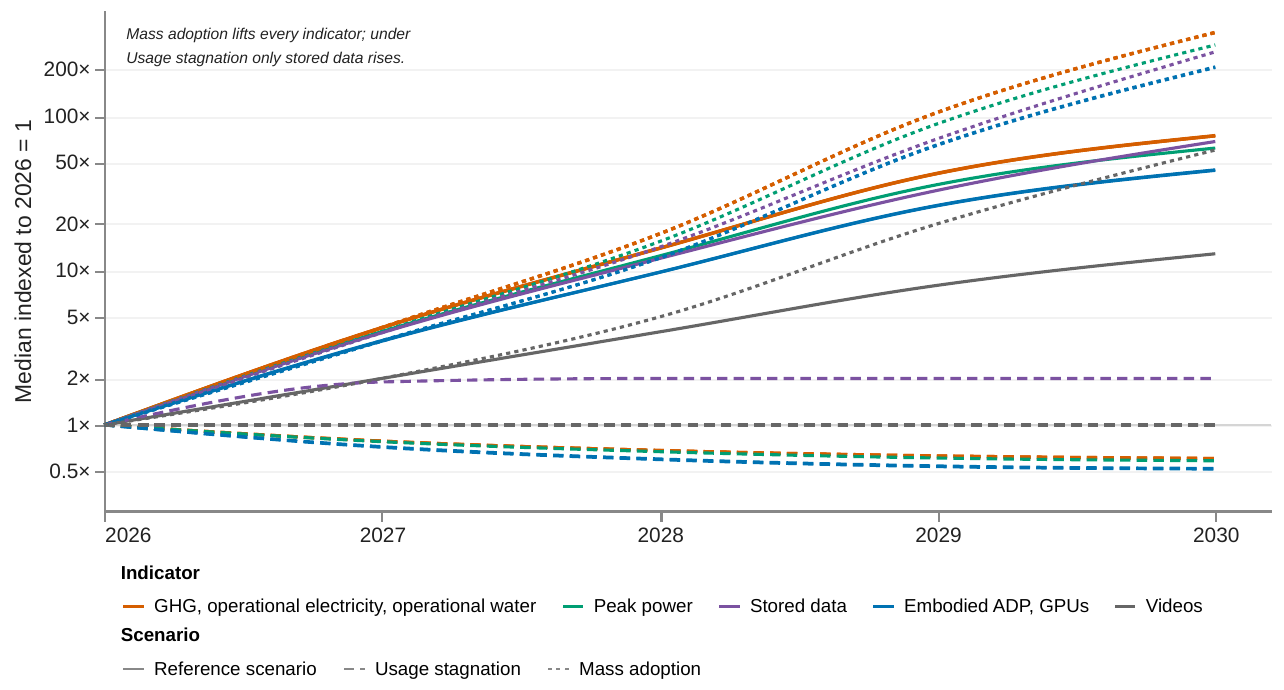}
    \caption{Divergence of the three scenarios, 2026--2030. Each line is the
        median of one indicator, indexed to its 2026 value. Colour identifies
        the indicator bundle and dash style identifies the scenario. Solid:
        \emph{Reference scenario}. Dashed: \emph{Usage stagnation}. Dotted:
        \emph{Mass adoption}.
    }
    \label{fig:scenario-divergence}
\end{figure}

\begin{figure}
    \centering
    \includegraphics[width=1\linewidth]{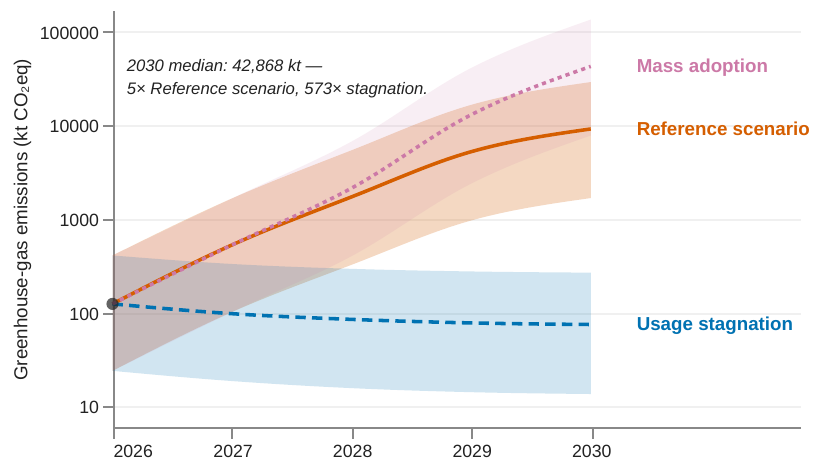}
    \caption{Greenhouse-gas trajectories under the three scenarios, 2026--2030.
        Lines are medians and bands span p5--p95 over 10{,}000 Monte Carlo
        samples; all three scenarios share the same 2026 distribution.}
    \label{fig:impacts-distributions-2026-2030}
\end{figure}

The scenarios differ in how demand, video quality and training energy
evolve:
the \emph{Reference scenario}, in which video volume, resolution and duration all
grow and the energy of each training run grows by 15\% a year;
\emph{Usage stagnation}, which freezes
volume, resolution and duration at their 2026 values and holds training energy
constant; and \emph{Mass adoption}, which shares the technology and quality
trajectories of the \emph{Reference scenario} but assumes a demand path far
above it (detailed below).
\autoref{fig:scenario-divergence} and \autoref{fig:impacts-distributions-2026-2030} trace the three scenarios over the period.

The volume trajectory of the \emph{Reference scenario} is deliberately modest:
video volume doubles in each of 2027, 2028 and 2029, then grows by 60\% in
2030, for a cumulative $12.8\times$ over four years (average yearly multiple
about $1.9$), which stays well below the growth Google reported for its AI
products, whose monthly processed tokens rose from 9.7 trillion in May 2024 to
more than 3.2 quadrillion in May 2026~\cite{pichaiGoogleIO2026}.
The scenario is therefore a lower bound on demand rather than a central estimate.
It remains plausible because video serving is expensive enough to give
providers a brake that text and image generation arguably lacked: OpenAI
discontinued Sora in 2026~\cite{openaiSoraDiscontinuation2026}, showing that
usage surges in this modality can contract quickly.

\textbf{Impact analysis in 2030 by scenario.}
Under the \emph{Reference scenario}, the number of videos grows 12.8$\times$ between
2026 and 2030, but electricity grows 74.1$\times$ and GHG emissions 74.0$\times$, nearly six times as fast as demand.
The gap comes from video quality rather than volume: energy per video rises
from 24.4 to 214~Wh ($8.8\times$) as resolution and duration increase, in line with the
near-quadratic scaling reported for
current T2V models~\cite{delavandeVideoKilledEnergy2025,jeghamLightsCameraCarbon2026}.
This figure illustrates two opposing forces: quality growth alone (more pixels, longer clips) would have driven energy per video up
$20.9\times$ by 2030; the modelled technology efficiency trajectory mitigates this to the observed $8.8\times$ factor.

The same mechanism shows up in the infrastructure.
Effective latency rises $6.15\times$, so a fleet $44.3\times$ larger serves only 12.8$\times$ more videos.
Indicators also diverge as they grow: ADP rises $44.7\times$ against
$74.0\times$ for GHG emissions, because ADP tracks fleet size whereas
carbon tracks both fleet size and per-video energy.
In absolute terms, 2030 sees 53{,}611~GWh consumed, 8{,}084~MW of peak power, and
9{,}202~kt CO\textsubscript{2}eq emitted --- on the same per-capita distribution of the planetary
climate boundary, the annual
emissions of about 9.3 million people~\cite{salaEnvironmentalSustainability2020}.

Under \emph{Usage stagnation}, with demand, resolution and duration frozen, impacts
still fall by about two-fifths by 2030 (GHG $\times0.60$, electricity $\times0.61$):
energy per video falls to 10.2~Wh ($\times0.42$), but the fleet keeps about
half of its 2026 size ($\times0.52$), because the training GPUs, whose
count does not depend on output volume, offset the shrinking inference fleet.
By 2030 training also dominates this shrinking footprint (about half of its
GHG emissions and electricity), so under frozen demand the remaining footprint is
set by the training race.
Storage is the only quantity that still grows in this scenario ($\times2.0$),
since retained data keeps accumulating towards the retention-window steady state
even at constant output.
This two-fifths reduction is the most favourable outcome in our scenarios,
and it holds only while usage, resolution and duration stay frozen at their
2026 levels; the Discussion weighs how likely that condition is.

The \emph{Reference scenario} is a lower bound on demand, so a third
scenario, \emph{Mass adoption}, tests how the 2030 impacts respond to a
demand path above it: video volume doubles in 2027, grows $2.5\times$ in
2028, $4\times$ in 2029 and $3\times$ in 2030, a cumulative $60\times$
against $12.8\times$ for the \emph{Reference scenario}, with technology and
quality trajectories shared, so energy per video is identical and every
impact scales with volume.
In 2030, \emph{Mass adoption} reaches 429.8 billion videos, 249{,}640~GWh of
electricity, and 42.9~Mt CO\textsubscript{2}eq of emissions (5th--95th
percentiles: 7.8--135.7~Mt), with a GPU fleet of 14.9 million units and a peak
power of 37.5~GW (\autoref{supp:tab:scenarios-2030}); on the per-capita
distribution of the planetary climate boundary, these emissions correspond
to those of about 44 million people~\cite{salaEnvironmentalSustainability2020}.
We read the scenario as an illustrative sensitivity case rather than a
forecast: whether such accelerating rates transfer to video depends on forces
the model does not represent, such as supply constraints, pricing, and
provider strategy, which the recent Sora discontinuation
illustrates~\cite{openaiSoraDiscontinuation2026}.

\textbf{Required optimization rate (inversion analysis).} We finally invert the model: under the \emph{Reference scenario}, even a sustained 30\% efficiency improvement per year, the fastest rate we test, leaves 2030 GHG emissions 27.5$\times$ above their 2026 level. Holding emissions flat at their 2026 level under that scenario's demand would instead require improving the inference stack by about 75\% per year compounded, about seven times the roughly 10\% per year our technology trajectory models. 
Holding emissions flat would drive the optimization factor from 0.279 in 2026 down to about 0.0009 by 2030, below the best efficiency measured for a production model (Veo~3, at 0.084~\cite{jeghamLightsCameraCarbon2026}) in every year from 2027, so no software improvement of known magnitude can hold emissions flat.
Software optimization alone therefore cannot hold the footprint steady; the remaining levers act on demand or on training. %
In short, at every optimization rate we test, efficiency gains alone leave
the 2030 footprint far above its 2026 level, so within our scenarios the
trajectory bends only when demand or video quality stop growing.

\section{Discussion}\label{sec:discussion}

Our scenarios show that the trajectory of text-to-video generation over
the next five years is set jointly by video volume, quality, and hardware
efficiency: under the \emph{Reference scenario}, GHG emissions grow almost six
times as fast as the number of videos (74.0$\times$ against 12.8$\times$); under \emph{Usage stagnation}, the same
technological levers, with demand held fixed, trim the footprint by about two-fifths
instead; and under \emph{Mass adoption}, with demand above the \emph{Reference}
path, the 2030 impacts scale with volume.
In this comparison, demand growth is the driver and quality growth the
mechanism: quality sets how fast the energy of one video rises, and demand
decides how many videos carry that energy.

\subsection{Demand outweighs efficiency}
The 2030 outcomes under the \emph{Reference scenario} and \emph{Usage
stagnation} are more than a hundredfold apart in GHG emissions.
Within the uncertainty we represent, the industry's trajectory
still separates the outcomes, though by a margin comparable to the spread of
the model's own uncertain inputs.

Neither trajectory is neutral against climate goals.
The digital sector's own target (Recommendation ITU-T L.1470, aligned with the Paris
Agreement~\cite{ITUL1470}) calls for cutting emissions by 45\% between 2020 and 2030.
Our \emph{Reference scenario} runs in the opposite direction,
multiplying operational electricity and GHG emissions nearly seventy-five-fold within four
years.
Sector-level studies reach the same direction: the Shift Project finds that even with grid decarbonisation, data-centre emissions keep growing at about 9\% per year, and it derives absolute caps on data-centre electricity use~\cite{IntelligenceArtificielleDonnees}.
The model already includes the main technical levers the sector proposes
(fleet renewal and inference-stack optimisation) and growth still
multiplies impacts, so the gap with goal trajectories follows from the growth
scenario itself rather than from our choices.

\emph{Usage stagnation} shows the right direction of travel: freezing volume
and video quality while hardware improves would cut the footprint by about
two-fifths, a direct reduction of the sector's total impacts %
Whether such a freeze is achievable is an economic question: across
semiconductors, mobile access and data centres, efficiency improvements have
been absorbed by rising technology affluence~\cite{bolMooresLawICT2021}, and
at the global scale by growing
demand~\cite{alpineAIdrivenProductivityGains2026, luccioniEfficiencyGainsRebound2025}.
The comparison that matters is therefore between efficiency gains and demand
growth. Under the \emph{Reference scenario}, holding emissions flat would require
an optimization factor below the best known for a production model. This suggests
that software optimization alone cannot hold emissions flat under this scenario.

The \emph{Mass adoption} scenario makes the demand side of this comparison
concrete. Its volume path grows between two and four times per year, within
the expansion that large providers reported for text- and image-generation
services over 2024--2026~\cite{pichaiGoogleIO2026}, yet it multiplies the 2030
footprint nearly fivefold beyond the \emph{Reference} path; because the two
scenarios share their technology, quality and training trajectories, that
additional footprint traces to demand alone. As a sensitivity case, the scenario isolates demand by design: technology, quality and training are held identical, so the
sustainability question turns on demand as much as on technology efficiency.

\subsection{Growth shifts burden to hardware}

Longitudinal estimates show machine-learning training impacts growing
exponentially through 2025 despite major hardware and algorithmic gains, and even under simulated carbon-aware training, a rebound attributed to ever larger models~\cite{morandEnvironmentalImpactsMachine2025}, while the embodied footprint of producing AI GPUs keeps rising steeply in parallel~\cite{morandRisingUnsustainabilityAI2026}.
Efficiency on an expanding infrastructure therefore does not remove the
burden; it shifts it toward the embodied cost of producing new hardware and toward rebound-driven growth in demand.

Under \emph{Mass adoption}, video volume grows by a cumulative $60\times$ over
2026--2030 and the GPU fleet grows with it, so the material footprint of
manufacturing that hardware would reach levels far beyond those observed
today; supply constraints of that kind are not represented in the model.
Supply-chain evidence indicates where such constraints would bite first: in
2025, the four largest AI chip designers (NVIDIA, Google, AMD, and Amazon)
consumed about 90\% of global advanced-packaging capacity and of
high-bandwidth-memory supply, but only about 12\% of advanced logic-die
production~\cite{EpochChipSupplyChain2026}.

Because those four designers already take up nearly all packaging and memory
supply, sustaining hardware growth on the scale of the \emph{Mass adoption} path would
rest on expanding capacity that AI demand already dominates, and that
expansion would proceed through new manufacturing facilities on multi-year
lead times; advanced logic, where AI demand held a smaller share in 2025,
looks like a softer constraint so far, though the industry's move to 3~nm
processes is likely to raise that share considerably.

\subsection{Estimates are lower bounds}

Sizing the trajectory the industry follows is difficult because
usage is opaque: to our knowledge, only a handful of providers have published rough generation counts, our own measurements cover two more platforms, and every other line of evidence is disclosed at the
provider's convenience, over ill-matched time windows.
Disclosure of inference conditions (data centres used by each
generation, power-usage effectiveness, and the hardware fleet) would
let future estimates replace these triangular priors with measured,
location-weighted values.
It would also make generation location visible, which grid operators and regulators need.

Our estimates should be read as a lower bound, for several reasons.
The model holds the generation architecture fixed across 2026--2030, while real models
are likely to grow more capable and more expensive to run over the same
period.
The scope excludes the diffusion of videos to viewers (content
delivery, playback, and social-media storage), and audio generation remains
excluded.
Text-to-video is also only one usage among many; our results describe it in
isolation and should not be read as a proxy for the footprint of AI as a whole.

Training is the second-largest component of the 2026 footprint: 31\% of the
summed GHG component medians and about one-third of operational electricity.
That share, however, reflects what is publicly known rather than what
training actually costs: it follows from the assumed retraining cadence
(20--40 runs per year in our prior) and the assumed per-run energy, not from
a measurement.
No provider we study discloses the compute or energy behind its closed-source
model, so our training-cost model is calibrated on a few published runs of
smaller, open-source text-to-video models, scaled by a development overhead.
Spending data point in the same direction: at three AI companies where the
split is published or estimated, compute for research and development alone
takes 40--52\% of total operating expenses~\cite{EpochCompanySpendingBreakdown2026}, which is consistent with
training-related compute being larger than documented runs suggest.

Other simplifications shape how the results should be read.
The GPU fleet is treated as dedicated to inference, with no sharing across
services; the generation model is a black box with no GPU-memory accounting; one GPU serves one
inference at a time, with no credit for batching (per-video energy varies within only
3.3\% across batch sizes 1 to~4~\cite{jeghamLightsCameraCarbon2026});
and server overhead is applied uniformly across GPU generations despite being
measured on a single reference architecture.

We deliberately excluded several adjacent effects from the model.
Following Kaack \etal, our estimates cover the computing-related impacts of machine learning, quantified operationally and at the embodied level for training, inference, and storage, while the immediate impacts of the applications it enables and the system-level impacts of their wider use lie outside our boundary~\cite{kaackAligningArtificialIntelligence2022}.
Rebound and substitution effects belong to those wider, indirect categories.

\subsection{Effects outside the boundary}

A more efficient pipeline could lower the per-video impacts we
compute, but also lower the barrier to generating and drive total
demand up faster than efficiency brings it down, a rebound our scenarios do
not capture~\cite{luccioniEfficiencyGainsRebound2025}.
Provider strategies point the same way: efficiency is deployed to enable
larger activity volumes, an ``intended enablement'' documented for 5G and
model training rather than measured for video
generation~\cite{morandWhenReboundEffect2026}.
Likewise, AI video generation that takes the place of traditional filming
displaces burdens that sit
outside our boundary (studios and sets, location generators, travel by cast,
crew and equipment), about 2,840~t~CO\textsubscript{2}e for an average large-budget film production~\cite{arupScreenNewDeal2020}.
Replacement, however, is not the only outcome: generative clips are often
added to footage shot conventionally rather than replacing it, as when Netflix reported hundreds of titles using generative-AI workflows in post-production~\cite{netflixGenAIWorkflows2026}; the
traditional burdens then persist, and the AI burden comes on top instead of
displacing them.
We do not evaluate that balance, and substitution carries a social
counterpart (creative-sector employment, already a contested point in
professional animation production~\cite{luminateAnimationAI2025}) that we
equally leave unquantified.

Economy-wide modelling reinforces this caution: in a global energy--economy
model, AI productivity gains raised rather than lowered fossil-fuel
CO\textsubscript{2} emissions~\cite{alpineAIdrivenProductivityGains2026}.
Read against this record, the \emph{Reference scenario} and \emph{Mass adoption} are the less
conditional of the three: the roughly two-fifths reduction under \emph{Usage
stagnation} holds only while usage stays frozen, a condition the
economy-wide evidence does not support.
We leave the direction and size of these effects, and of any regulatory
lever able to shape them, to future work grounded in usage data that does
not yet exist.

\section{Methods}\label{sec:methods}

In this section, we present how we estimate the yearly resource needs and environmental impacts of a text-to-video generation service.
The service represents the combined output of existing providers, rather than one user's individual generation.
The model follows the service from model training to inference and storage.
Detailed equations, parameters, sources, and implementation rules appear in the Supplementary.

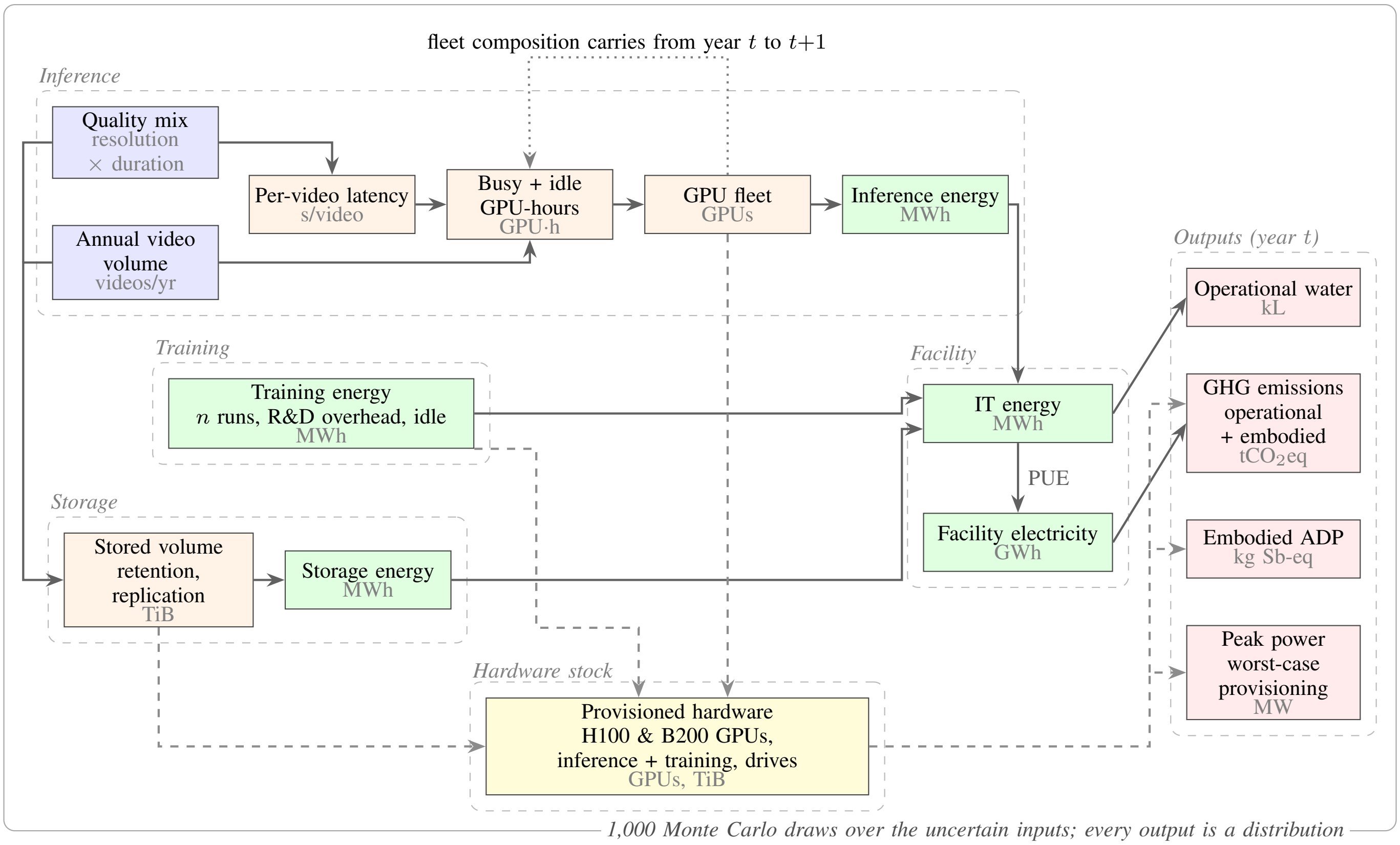
\begin{figure*}[t]
\centering
    \begin{tikzpicture}[node distance=6.5mm and 4mm]

    \node[inp] (qmix) at (0,0.8) {Quality mix\\[-2pt]{\fnunit resolution $\times$ duration}};
    \node[inp] (demand) at (0,-0.75) {Annual video volume\\[-2pt]{\fnunit videos/yr}};
    \node[calcnode] (lat) at (2.54,0) {Per-video latency\\[-2pt]{\fnunit s/video}};
    \node[calcnode, right=of lat] (hours) {Busy + idle GPU-hours\\[-2pt]{\fnunit GPU$\cdot$h}};
    \node[calcnode, right=of hours] (fleet) {GPU fleet\\[-2pt]{\fnunit GPUs}};
    \node[pwr, right=of fleet] (einf) {Inference energy\\[-2pt]{\fnunit MWh}};

    \draw[flow] (qmix.east) -| (lat.north);
    \draw[flow] (demand.east) -| (hours.south);
    \draw[flow] (lat) -- (hours);
    \draw[flow] (hours) -- (fleet);
    \draw[flow] (fleet) -- (einf);

    \node[pwr, text width=3.8cm] (train) at (2.4,-2.7)
        {Training energy\\ $n$ runs, R\&D overhead, idle\\[-2pt]{\fnunit MWh}};

    \node[calcnode, text width=2.3cm] (svol) at (0.3,-4.85)
        {Stored volume\\ retention, replication\\[-2pt]{\fnunit TiB}};
    \node[pwr, right=of svol] (estor) {Storage energy\\[-2pt]{\fnunit MWh}};

    \draw[flow] (svol) -- (estor);
    \coordinate (lext) at (-1.45,0);
    \draw[thick, black!62] (qmix.west) -- (-1.45,0.8) -- (-1.45,-0.75);
    \draw[flow] (demand.west) -- (-1.45,-0.75) |- (svol.west);

    \node[pwr, text width=2.3cm] (EIT) at (11.4,-2.7)
        {IT energy\\[-2pt]{\fnunit MWh}};
    \node[pwr, text width=2.3cm, below=9mm of EIT] (E)
        {Facility electricity\\[-2pt]{\fnunit GWh}};

    \draw[flow] (einf) -| (EIT);
    \draw[flow] (train.east) -- (9.9,-2.7) |- ([yshift=2mm]EIT.west);
    \draw[flow] (estor.east) -- (9.9,-4.85) |- ([yshift=-2mm]EIT.west);
    \draw[flow] (EIT.south) -- (E.north) node[midway, right, font=\footnotesize] {PUE};

    \node[hwn, text width=4.8cm] (hw) at (7.0,-7.0)
        {Provisioned hardware\\ H100 \& B200 GPUs, inference + training, drives\\[-2pt]{\fnunit GPUs, TiB}};

    \coordinate (bus) at ([yshift=9mm]hw.north);

    \draw[hwflow] (fleet.south) -- (fleet.south |- hw.north);
    \draw[hwflow] (train.south east) -- ([xshift=8mm]train.south east)
        -- ([xshift=8mm]train.south east |- bus) -- (6.5,0 |- bus) -- (6.5,0 |- hw.north);
    \draw[hwflow] (svol.south) |- (hw.west);

    \node[outp] (oW) at (14.7,-1.2) {Operational water\\[-2pt]{\fnunit kL}};
    \node[outp, below=6mm of oW] (oG) {GHG emissions\\ operational + embodied\\[-2pt]{\fnunit tCO$_2$eq}};
    \node[outp, below=6mm of oG] (oM) {Embodied ADP\\[-2pt]{\fnunit kg Sb-eq}};
    \node[outp, below=6mm of oM] (oP) {Peak power\\ worst-case provisioning\\[-2pt]{\fnunit MW}};

    \draw[flow] (EIT.east) -- (oW.west);
    \draw[flow] (E.east) -- (oG.west);

    \coordinate (cG) at (13.10,0 |- oG.west);
    \coordinate (cM) at (13.10,0 |- oM.west);
    \coordinate (cP) at (13.10,0 |- oP.west);
    \draw[hwflow] (hw.east) -| (cP) -- (oP.west);
    \draw[hwflow] (cP) -- (cM) -- (oM.west);
    \draw[hwflow] (cM) -- ([yshift=2.5mm]cG) -- ([yshift=2.5mm]oG.west);

    \draw[flow, dotted, gray] (fleet.north) |- (5.08,1.9) -- (hours.north);
    \node[font=\footnotesize] (rectxt) at (6.35,2.08) {fleet composition carries from year $t$ to $t{+}1$};

    \begin{pgfonlayer}{background}
        \node[lane, fit=(qmix) (demand) (lat) (hours) (fleet) (einf)] (flInf) {};
        \node[lane, fit=(train)] (flTr) {};
        \node[lane, fit=(svol) (estor)] (flSt) {};
        \node[lane, fit=(EIT) (E)] (flFac) {};
        \node[lane, fit=(hw)] (flHw) {};
        \node[lane, fit=(oW) (oG) (oM) (oP)] (flOut) {};
    \end{pgfonlayer}

    \node[lname] at (flInf.north west) {Inference};
    \node[lname] at (flTr.north west) {Training};
    \node[lname] at (flSt.north west) {Storage};
    \node[lname] at (flFac.north west) {Facility};
    \node[lname] at (flHw.north west) {Hardware stock};
    \node[lname] at (flOut.north west) {Outputs (year $t$)};

    \begin{pgfonlayer}{background}
        \node[outer, fit=(flInf) (flTr) (flSt) (flFac) (flHw) (flOut) (rectxt) (lext)] (mc) {};
    \end{pgfonlayer}
    \node[anchor=east, fill=white, inner sep=2pt, font=\itshape\footnotesize, text=black!70]
        at ([xshift=-6mm]mc.south east)
        {1{,}000 Monte Carlo draws over the uncertain inputs; every output is a distribution};

    \end{tikzpicture}
\caption{Model pipeline for one simulated year $t$. Solid arrows carry energy
and volume quantities, dashed arrows carry the hardware stock behind the embodied
impacts, and the dotted arrow shows the year-to-year recursion. 
}
\label{fig:model-pipeline}
\end{figure*}

\subsection{Scope and study boundaries}\label{subsec:scope}

The scope of our study is the deployment of a text-to-video generation service.
It includes training and retraining, inference, and the storage of generated videos.
Training and retraining cover the sector's model-development activity: each laboratory that trains AI video-generation models contributes one to two trainings per year, so the model draws 20 to 40 retraining runs per year in total (Supplementary Methods~\ref{supp:retraining}).
It excludes content delivery, video playback, and storage on social-media platforms.
It also excludes audio generation, which a small share of newer models performs alongside the video.
We report operational electricity and operational water as operational impacts. The water
metric is the facility water input per unit of IT energy, from a data-centre
WUE: direct potable and non-potable water, with indirect water excluded.
We report greenhouse-gas emissions as operational and embodied impacts.
We report the embodied abiotic depletion potential (ADP) for the embodied phases only as an embodied impact only.

\subsection{Model overview}\label{subsec:model_structure}

\autoref{fig:model-pipeline} shows the model's structure for one simulated
year. The model runs year by year: each year, it combines yearly inputs into
per-video latencies, fleet and energy quantities, and yearly impacts, and
passes the hardware stock to the following year.
Each step is detailed below.

The model first estimates per-video latency from video quality, duration, frame rate, and GPU model, following the compute-bound DiT latency model of Delavande \etal~\cite{delavandeVideoKilledEnergy2025}.
It weights these latencies by the annual resolution and duration mix and the GPU fleet mix.
It then converts the resulting workload into active and idle GPU-hours, energy, and required GPUs.
Training adds a per-run energy term and a development overhead.
Storage adds the retained data volume, replication, and storage energy.
The model applies server, IT-equipment, and data-centre overheads before it derives facility electricity, water, and operational emissions.
It amortises embodied GPU, storage, and training impacts over the relevant hardware lifetime.
The model passes the fleet and stored data from year $t$ to year $t+1$.

The latency model is fitted on measurements of WAN2.1-T2V-1.3B on a single H100, and we transfer it to other GPUs by their peak-throughput ratio (Supplementary Methods).
The fitted latency therefore describes an unoptimized research configuration, not a deployed endpoint. Deployed services use kernel optimization, distillation, quantization, caching, and similar techniques to accelerate generation, so we apply one inference-optimization factor $\eta_{\text{opt}}$ that reduces the fitted latency and the resulting energy.
We calibrate $\eta_{\text{opt}}$ against the per-clip energy that Jegham \etal~\cite{jeghamLightsCameraCarbon2026} report for Seedance~1.5, on our reference workload of one 8-second, 720p clip on a single H100; this yields $\eta_{\text{opt}}=0.279$ for the base year.

\subsection{Annual volume}\label{subsec:volumetry}

We estimate annual video volume from two platform measurements and provider disclosures.
We measure public Sora output from its public feed and HeyGen's text-to-video output from its homepage counter.
Provider disclosures provide an independent scale check for services outside these two platforms.
The Supplementary give the crawl windows, counter processing, disclosure treatment, and utilisation calculation.

The model uses the Sora resolution and duration distributions to represent a generic short-clip service in 2026.

\subsection{Parameters and data sources}\label{subsec:variables}

The principal inputs of the model are annual video volume, video quality, GPU fleet composition, GPU utilisation, inference optimisation, training energy, storage retention, replication, PUE, WUE, and grid carbon intensity.
We model H100 and B200 GPUs in an 8-GPU node configuration.
Each GPU model is characterised by its rated power (TDP), its measured idle draw, and its embodied impacts, which include an amortised share of the host server.
We report embodied resource use as the abiotic depletion potential (ADP), which aggregates the mineral and the fossil-resource components of the underlying hardware datasets.
The model amortises these embodied impacts over the GPU lifespan as the fleet renews year by year.
Storage hardware is modelled from the stored volume: a per-video file size fitted on our Sora scrape, accumulated over a rolling retention window and multiplied by a replication factor, then charged with an operational energy per unit of stored data and embodied impacts per unit of storage hardware.
The Supplementary list all input and output variables, their units, distributions, sources, and fixed values.
The replication package contains the source data and scenario definitions.

\subsection{Uncertainty and scenarios}\label{subsec:uq}

We represent uncertain numeric inputs with probability distributions and draw one coherent sample for each field in each Monte Carlo run.
The final analysis uses 10,000 samples and seed 42 for the three scenarios.
We also run a one-at-a-time sensitivity analysis with the same seed and sample count.

All scenarios share the same technology trajectory.
The \emph{Reference scenario} increases video volume, resolution, and duration.
The \emph{Usage stagnation} scenario freezes volume and video quality at their 2026 values.
The \emph{Mass adoption} scenario applies a faster volume path (yearly multiples of 2, 2.5, 4, and 3) with the same quality and technology trajectories.
All scenarios renew the inference fleet and improve utilisation and inference optimisation.

Three assumptions limit the model.
It treats generation models as black boxes without explicit GPU-memory requirements.
It assigns one inference to one GPU and gives no batching credit.
It represents training energy and training GPU needs without an explicit training purchase schedule or bursty concurrency.
These choices make the estimates a lower bound for real deployments of this service.

\subsection{Limitations}\label{subsec:limits}

The estimates are a lower bound for the specialized text-to-video
ecosystem. The scope excludes audio generation, the diffusion
of videos to viewers, and generation usages outside specialized T2V
services, and it holds the generation model's size and inherent energy
cost constant over the simulated years. Each of these scope decisions
removes real demand or future growth from the estimate, so each one
pushes the results down rather than up.

Several inputs describe the market rather than measure it. The
resolution and duration mixes, and the storage file-size model, come
from a single platform (Sora) and may not represent the clips generated
worldwide. The fleet is restricted to two GPU models, H100 and B200, in
an 8-GPU node configuration. Facility overheads, water use, and grid
carbon intensity are drawn from European distributions and are not
weighted by where generations actually run; for carbon intensity, the
location of most generation remains an open question.

Training is the least measured part of the model. It is calibrated on a
few published runs of small open-source models scaled by a development
overhead, because no provider we study discloses the compute behind its
closed-source model. The model also spreads each year's training
GPU-hours uniformly across the year; this keeps the provisioned fleet
busy at all times and needs fewer GPUs, at lower peak power, than a real
campaign that concentrates the same work into weeks. Finally, the
generation model's size and training time do not grow over the years, so
the increase of the annual training cost is artificially low. All of
these choices push the training estimates down.

Three further calibrations rest on a single reported data point each
: the inference-optimization factor, the per-run training energy, and
the transfer of latency from H100 to B200 GPUs. The embodied GPU figures
additionally carry at least a factor-of-two uncertainty, as two public
data sources disagree by that margin. These inputs would benefit from
replacement by measured distributions as more data becomes public.

\section*{Supplementary information}

The full numerical inventories appear in the Supplementary Results, and the detailed equations, parameter tables, data sources, and uncertainty procedures in the Supplementary Methods.

\section*{Acknowledgements}
This research was conducted with the support of the Observatory on AI’s Environmental Footprint led by the AI and Society Institute at ENS.

\section*{Author contributions}
T.C. designed the model and wrote the code; A.B., G.G. and A.-L.L. supervised the work. All authors reviewed the manuscript.

\section*{Competing interests}
The authors declare no competing interests.

\section*{Data availability}
The model inputs, archived simulation results and crawl data are available in the replication package at \url{https://doi.org/10.5281/zenodo.24163581}.

\section*{Code availability}
The simulator code and analysis notebooks are available in the replication package at \url{https://doi.org/10.5281/zenodo.24163581}.

\bibliographystyle{IEEEtran}
\bibliography{biblio}%

\newpage

\renewcommand{\thesection}{S\arabic{section}}
\renewcommand{\thesubsection}{\thesection.\arabic{subsection}}
\renewcommand{\thesubsectiondis}{\thesubsection.}
\renewcommand{\theequation}{S\arabic{equation}}
\renewcommand{\thetable}{S\arabic{table}}
\renewcommand{\thefigure}{S\arabic{figure}}
\setcounter{section}{0}
\setcounter{subsection}{0}
\setcounter{equation}{0}
\setcounter{table}{0}
\setcounter{figure}{0}

\section{Supplementary Results}\label{supp:results}

This part of the paper gives the numerical details behind the figures of the
Results section.
All intervals correspond the 5th and 95th percentiles over 10{,}000 Monte Carlo
samples of the seeded reference run.

\subsection{Full 2026 inventory}\label{supp:inventory}

\autoref{supp:tab:impacts-2026}  lists all quantities reported by the model for 2026, the first year of the simulation.
It complements \autoref{fig:impacts-2026}
The 2026 values are identical under all three
scenarios, because the
scenarios diverge only after 2026.

\begin{table}[h]
    \centering
    \resizebox{\columnwidth}{!}{
        \begin{tabular*}{\linewidth}{@{\extracolsep{\fill}}l|cr}
            \toprule
                                      & Unit                     & Median [p5--p95]       \\
            \midrule\addlinespace[2.5pt]
            \multicolumn{3}{@{}l}{\emph{Demand input}}\\[2pt]
            Number of videos          & billion videos           & 7.16 [6.28--8.79]     \\
            \midrule\addlinespace[2.5pt]
            \multicolumn{3}{@{}l}{\emph{Environmental outputs}}\\[2pt]
            GHG Emissions             & kt CO\textsubscript{2}eq & 124.3 [23.9--409.4]     \\
            Operational electricity used & GWh                   & 724 [477--1{,}161]     \\
            Operational water consumed& ML                       & 343 [237--526]         \\
            Embodied ADP              & kg Sb-eq                 & 860 [515--1{,}602]     \\
            \midrule\addlinespace[2.5pt]
            \multicolumn{3}{@{}l}{\emph{Infrastructure outputs}}\\[2pt]
            Peak Power                & MW                       & 130.8 [84.1--215.1]       \\
            GPUs Required             & k GPUs                   & 72.6 [52.2--110.0]       \\
            Stored Data (AI provider) & PiB                      & 48.4 [42.4--59.3]      \\
            \midrule\addlinespace[2.5pt]
            \multicolumn{3}{@{}l}{\emph{Per-video unit factors}}\\[2pt]
            Video Size        & MiB/video                & 2.42                   \\
            Energy per Video          & Wh/video                 & 24.4                   \\
            Effective Latency         & s/video                  & 125                    \\
            \bottomrule
        \end{tabular*}
    }
    \begin{minipage}{\linewidth}
        \vspace{1mm}
        \footnotesize
        Median and 5th--95th percentile over 10{,}000 Monte Carlo samples.
        The three per-video quantities at the bottom are deterministic
        functions of the fixed inputs (the resolution and duration mix, the
        hardware model, and the optimization factor), so they carry no
        sampled spread; they are reported as plain values.
        Energy per Video is the GPU-active term only, excluding idle draw,
        facility overheads, training, and storage.\\
    \end{minipage}
    \caption{Full 2026 inventory of the demand input, the environmental and
        infrastructure outputs, and the per-video unit factors that the model
        reports for text-to-video generation.}
    \label{supp:tab:impacts-2026}
\end{table}

\subsection{The 2026-2030 scenarios}\label{supp:scenarios}

\autoref{supp:tab:scenarios-2030} gives the 2030 medians of the three
scenarios
beside their common 2026 starting point, with the change factor for each.

The retraining count is held at the prior of Supplementary Methods~\ref{supp:retraining} in all three
scenarios; as a diagnostic sensitivity of this training-constancy assumption, a
stagnation variant with the count halved to $n\sim\text{uniform}(10,20)$ in
every year brings the 2030 median GHG emissions down from 74.9 to
55.2~kt CO\textsubscript{2}eq, a reduction of about 47\% against its own 2026 level instead
of about two-fifths (the variant is included in our replication
package).\label{supp:training-constancy}

\begin{table*}[t]
    \centering
    \resizebox{\textwidth}{!}{
        \begin{tabular}{@{\extracolsep{\fill}}l|c|r|rr|rr|rr}
            \toprule
                                      &                          &       & \multicolumn{2}{c|}{\emph{Reference scenario}} & \multicolumn{2}{c|}{\emph{Usage stagnation}} & \multicolumn{2}{c}{\emph{Mass adoption}}   \\
            \cmidrule(lr){4-5}\cmidrule(lr){6-7}\cmidrule(lr){8-9}
                                      & Unit                     & 2026  & 2030 & Change        & 2030 & Change        & 2030 & Change        \\
            \midrule\addlinespace[2.5pt]
            Number of videos          & billion videos           & 7.16  & 91.7  & $\times$ 12.8  & 7.16 & $\times$ 1.0    & 429.8 & $\times$ 60.0  \\
            \addlinespace[2.5pt]
            GHG Emissions             & kt CO\textsubscript{2}eq & 124.3  & 9{,}202 & $\times$ 74.0 & 74.9 & $\times$ 0.60  & 42{,}868 & $\times$ 345 \\
            Operational electricity used & GWh                   & 724   & 53{,}611 & $\times$ 74.1 & 439  & $\times$ 0.61  & 249{,}640 & $\times$ 345 \\
            Operational water consumed& ML                       & 343   & 25{,}567 & $\times$ 74.5 & 208  & $\times$ 0.61  & 119{,}026 & $\times$ 347 \\
            Embodied ADP              & kg Sb-eq                 & 860   & 38{,}393 & $\times$ 44.7 & 450  & $\times$ 0.52  & 177{,}899 & $\times$ 207 \\
            Peak Power                & MW                       & 130.8  & 8{,}084 & $\times$ 61.8 & 76.7 & $\times$ 0.59  & 37{,}538 & $\times$ 287 \\
            GPUs Required             & k GPUs                   & 72.6  & 3{,}218   & $\times$ 44.3 & 37.6 & $\times$ 0.52  & 14{,}895 & $\times$ 205 \\
            Stored Data (AI provider) & PiB                      & 48.4  & 3{,}295   & $\times$ 68.1 & 96.6 & $\times$ 2.0   & 12{,}520 & $\times$ 259 \\
            Average Video Size        & MiB/video                & 2.42  & 8.33  & $\times$ 3.45 & 2.42 & $\times$ 1.0    & 8.33  & $\times$ 3.45 \\
            Energy per Video          & Wh/video                 & 24.4  & 214   & $\times$ 8.8  & 10.2 & $\times$ 0.42   & 214   & $\times$ 8.8  \\
            Effective Latency         & s/video                  & 125   & 771   & $\times$ 6.15 & 36.8 & $\times$ 0.294  & 771   & $\times$ 6.15 \\
            \bottomrule
        \end{tabular}
    }
    \begin{minipage}{\linewidth}
        \vspace{1mm}
        \footnotesize
        Medians over 10{,}000 Monte Carlo samples; the 2026 column is shared by
        all three scenarios.
        Energy per Video is the GPU-active term only.\\
    \end{minipage}
    \caption{Demand, environmental, and infrastructure quantities of
        text-to-video generation in 2026 and in 2030 under the three scenarios.}
    \label{supp:tab:scenarios-2030}
\end{table*}

\subsection{Comparison with video-platform upload fluxes}\label{supp:flux-comparison}

To put the modelled volumes into perspective, we compare them with the
upload flux of established video platforms.

YouTube states that, on average, over 20 million videos are uploaded to
the platform daily, as of April
2025~\cite{youtubeTwentyYears2025}; the same figure is displayed on its
press page~\cite{youtubePress2025}.
This corresponds to roughly 7.3 billion videos per year, very close to the
2026 median of the model of 7.16 billion videos (about 20 million per day).
The closeness of the two counts is, however, a poor guide to volume, because
the two flows differ greatly in video length: the generated clips average
about 6.5 seconds in 2026 (76\% of them 5 seconds long, none longer than 20
seconds) and about 10.5 seconds in 2030, against a mean of 615 seconds
(median 126 seconds) in a random sample of YouTube
uploads~\cite{mcgradyDialingVideos2023}.
In seconds of footage, the unit we keep for the rest of this section, the
2026 median of the model is therefore about 1\% of the YouTube upload flux,
the 2030 Reference-scenario median about a fifth of it, and the
Mass-adoption median about equal to it.
For reference, the same comparison in raw video counts gives about 98\% of
this flux for 2026, thirteen times it for the 2030 Reference-scenario median
(about 251 million videos per day), and sixty times it for the 2030
Mass-adoption median (about 1.2 billion videos per day); counting videos
rather than seconds overstates the footage share by a factor of about 95 in
2026 and about 60 in 2030.

TikTok publishes no upload volumes; its transparency disclosures cover
moderation actions and user counts only.
The best available figure comes from a near-complete census of the
platform, which estimates 269 million posts created over one measured
24-hour window in April 2024~\cite{steelJustAnotherHour2026}, at a mean
post length of 20.7 seconds.
In seconds of footage, the 2026 median of the model is about 2\% of this
one-day flux, the 2030 Reference-scenario median about half of it, and the
Mass-adoption median about 2.2 times it.
For reference, the same comparison in raw video counts gives about 7.3\% of
this flux for 2026, about 93\% for the 2030 Reference-scenario median, and
more than four times the flux for the 2030 Mass-adoption median.

Three caveats bound these comparisons.
First, the platform figures are unaudited company statements or
sampling-based estimates with undisclosed definitions: YouTube does not
state whether private or unlisted uploads are counted, and the TikTok
figure is a one-day extrapolation that includes posts later deleted or
privatised (about 46\% of the posts sampled in the census were no longer
publicly accessible)~\cite{steelJustAnotherHour2026}.
Second, the footage ratios above inherit the uncertainty of the length
estimates on both sides.
On the model side, the 2026 mean comes from measured public Sora outputs,
but those outputs stand in for many providers whose clip lengths may
differ, and the 2030 mean follows the scenarios' assumed drift towards
longer videos.
On the platform side, the means come from single samples; the YouTube mean
is dominated by a minority of long uploads (the sample median is 126
seconds), so it is sensitive to sample composition, and the YouTube footage
flux is dominated by long-form content (Shorts, music, long-form uploads),
whereas the model counts generated short clips only.
Finally, seconds of footage measure volume only: one generated second and
one uploaded second are not comparable in production impact.
Third, the comparison is a flux-to-flux one; for stocks, YouTube reported
over 20 billion uploaded videos over its first twenty years as of April
2025~\cite{youtubeTwentyYears2025}, and a peer-reviewed census estimated
9.9 billion publicly visible videos in late
2022~\cite{mcgradyDialingVideos2023}.

This comparison has no consequence for the model itself: the volume prior
is anchored on provider disclosures and on the measured floor, not on
platform fluxes, so the two estimates are independent.
Its role is to bound the reading of the results.
In footage terms, the modelled volumes remain small shares of established
platform fluxes in 2026: about 1\% of the YouTube upload flux over a year,
and about 2\% of the TikTok one-day flux over a day.
Even the 2030 Mass-adoption median amounts to roughly one year of YouTube
uploads at the 2025 rate.

\subsection{Absolute one-at-a-time spreads}\label{supp:oat-absolute}

\autoref{fig:tornado-2026} normalises each output column to its longest bar, so
it ranks the inputs but hides the size of the spread they produce.
\autoref{supp:tab:oat-absolute} gives the same spreads in the unit of each
output.

\begin{table*}[t]
    \centering
    \resizebox{\textwidth}{!}{
        \begin{tabular}{@{\extracolsep{\fill}}l|rrrrr}
            \toprule
                                      & Required GPUs            & GHG emissions            & Operational electricity & Operational water & Embodied ADP \\
                                      & (k GPUs)                 & (kt CO\textsubscript{2}eq) & (GWh)       & (ML)  & (kg Sb-eq)   \\
            \midrule\addlinespace[2.5pt]
            Grid carbon intensity              & 0    & 340.9 & 0     & 0     & 0     \\
            Energy per training run            & 42.0 & 66.3  & 382.5 & 184.5 & 484.5 \\
            Data-centre energy overhead (PUE)  & 0    & 64.2  & 417.3 & 0     & 0     \\
            R\&D overhead on training          & 24.5 & 38.6  & 223.0 & 107.6 & 282.5 \\
            Number of videos generated         & 15.6 & 27.4  & 160.1 & 77.2  & 180.0 \\
            Retrainings per year               & 16.1 & 25.4  & 146.7 & 70.8  & 185.8 \\
            GPU lifespan                       & 0    & 12.2  & 0     & 0     & 793.9 \\
            GPU utilisation                    & 12.8 & 4.8   & 16.8  & 8.1   & 147.6 \\
            Data-centre water intensity (WUE)  & 0    & 0     & 0     & 134.4 & 0     \\
            \bottomrule
        \end{tabular}
    }
    \caption{Absolute p5--p95 output spread produced by each uncertain input
        alone, 2026, in the unit of each output (Reference scenario; 10{,}000
        Monte Carlo samples per input, all other inputs at their medians).
        Rows are sorted by decreasing GHG spread; a zero means the input does
        not affect that output. The video retention period produces no spread
        in any of the five outputs and is therefore omitted (it is also absent
        from \autoref{fig:tornado-2026}).}
    \label{supp:tab:oat-absolute}
\end{table*}

\subsection{Validation of our model against real-world data}

We finally confront the model's per-video predictions with market prices.

Our per-video cost is calibrated to the per-clip energy that Jegham \etal{}
report for Seedance~1.5, a mid-tier commercial service: 102.1~Wh for an 8-second,
720p clip on an H100~\cite{jeghamLightsCameraCarbon2026}.
Converting the model's GPU-seconds into an infrastructure cost at market rates and comparing with commercial per-clip prices checks whether the throughput level is plausible: at 2026 rates for H100-class hardware (0.0006--0.0022~USD per GPU-second
across on-demand, serverless, and contract offers~\cite{runpodPricing2026,fairstackCostBlog2026}),
that calibration point (about 525 GPU-seconds on an H100 at the
model's measured power draw) costs 0.3--1.2~USD of raw GPU infrastructure per clip and
falls inside the 0.072--1.50~USD band a cost-disclosing service charges
for its 5--8-second catalogue~\cite{fairstackCostBlog2026}.

\subsection{Sensitivity to the software-efficiency trajectory}\label{supp:opt-sensitivity}

The technology trajectory assumes that the inference-optimization factor falls
by a cumulative $\sim$35\% over 2026--2030, about 10\% per year compounded.
To test how the conclusions depend on that assumption, we rerun the
\emph{Reference scenario} deterministically, holding every input at its median
except the optimization factor, which follows a constant annual improvement
rate $g$ instead of the assumed decelerating profile.
\autoref{supp:tab:opt-sensitivity} reports the 2030 outcomes.

\begin{table}[h]
    \centering
    \begin{tabular}{@{\extracolsep{\fill}}l|rrr}
        \toprule
        Efficiency path & $\eta_{\text{opt}}$ in 2030 & GHG in 2030 & GHG growth \\
                        &                             & (kt CO\textsubscript{2}eq) & $\times$ 2026 \\
        \midrule\addlinespace[2.5pt]
        Assumed profile  & 0.181 & 8{,}957  & 73.4  \\
        None ($g=0\%$/yr) & 0.279 & 13{,}740 & 112.5 \\
        $g=5\%$/yr       & 0.227 & 11{,}207 & 91.8  \\
        $g=30\%$/yr      & 0.067 & 3{,}360  & 27.5  \\
        \bottomrule
    \end{tabular}
    \caption{Sensitivity of the \emph{Reference scenario} to the
        software-efficiency trajectory. Deterministic runs with all inputs at
        their medians; the inference-optimization factor either follows the
        assumed decelerating profile (cumulative $\sim$35\% over 2026--2030)
        or shrinks at a constant annual rate $g$.
        All runs share the 2026 starting point (122.1~kt CO\textsubscript{2}eq
        at $\eta_{\text{opt}}=0.279$), which differs from the Monte Carlo
        median quoted in the main text (124.3~kt) because a run with all
        inputs frozen at their medians does not reproduce the median of the
        Monte Carlo outputs.}
    \label{supp:tab:opt-sensitivity}
\end{table}

Across the tested range, the efficiency path moves the 2030 total by about
$4\times$ (13{,}740 vs 3{,}360~kt CO\textsubscript{2}eq), but no tested rate
approaches the roughly 75\% per year that holding 2030 emissions at their 2026
level would require.

\newcommand{\nrj}{C} %

\newcommand{\latency}{D} %
\newcommand{\gpuModel}{m}
\newcommand{\videoRes}{r}
\newcommand{\videoDur}{u}

\newcommand{\gpuMix}{\hat{\gpuModel}_i} %
\newcommand{\resMix}{\hat{\videoRes}_k} %
\newcommand{\durMix}{\hat{\videoDur}_j} %

\newcommand{\infSuffix}{{\text{inf}}} %
\newcommand{\trainSuffix}{{\text{train}}} %
\newcommand{\nbTrain}{n_\trainSuffix}
\newcommand{\nbGpuInf}{N_\infSuffix}
\newcommand{\nbGpuTrain}{N_\trainSuffix}
\newcommand{\tauInf}{\tau_\infSuffix}
\newcommand{\tauTrain}{\tau_\trainSuffix}
\newcommand{\powerRateTrain}{a_\trainSuffix}
\newcommand{\nrjPerTrain}{e_\trainSuffix}
\newcommand{\devOverhead}{k_{\text{dev}}}

\newcommand{\storSuffix}{{\text{stor}}} %

\newcommand{\trainModel}{\gpuModel_\trainSuffix}

\newcommand{\nrjInfAct}{\hat{\nrj}_\infSuffix}

\section{Supplementary Methods}\label{supp:methods}

This part of the paper gives the detailed model definition that supports the short Methods section above.
It provides the equations, parameter definitions, data sources, uncertainty procedure, and modelling boundaries.
We refer to \autoref{sec:methods} and \autoref{fig:model-pipeline} for an overview.

\subsection{Scope and impact inventory}\label{supp:scope}

The scope of the study is the \emph{deployment of a text-to-video generation
service}, sized to the combined generation volume of all existing services,
rather than the generation of a single video by a single user. %
It covers three cost components: training (the annual cost of $\nbTrain$ trainings
and re-trainings), inference (a cost proportional to the volume of
generations), and storage (the cumulative cost of retaining generated videos).
It excludes the diffusion of videos to viewers, \ie content delivery,
playback, social-media storage and sharing and storage on other platforms.

Audio generation is out of scope.
Most current text-to-video models produce silent video; only a growing share of newer models, sometimes called text-to-video-audio models, also synthesize an audio track alongside the video~\cite{jeghamLightsCameraCarbon2026}.
We exclude this audio component for clarity and simplicity.
The exclusion keeps our estimates conservative: any audio generated only adds to the impacts we report, so the figures remain a lower bound for services that generate audio as well as video.

Not every impact is estimated for every phase.
Greenhouse gas emissions cover both the operational and the embodied phase.
Operational electricity and operational water are operational only.
Embodied abiotic depletion potential (ADP, kg Sb-eq) is embodied only.
\autoref{tab:supp-impact-boundary} summarises which phases each reported indicator covers.

\begin{table}[t]
    \centering
    \begin{tabular}{@{}lcccccc@{}}
        \toprule
        & \multicolumn{2}{c}{Training} & \multicolumn{2}{c}{Inference} & \multicolumn{2}{c}{Storage} \\
        \cmidrule(lr){2-3} \cmidrule(lr){4-5} \cmidrule(lr){6-7}
        Indicator & op. & emb. & op. & emb. & op. & emb. \\
        \midrule
        Operational electricity & \cmark & -- & \cmark & -- & \cmark & -- \\
        Operational water       & \cmark & -- & \cmark & -- & \cmark & -- \\
        GHG (t CO$_2$eq)        & \cmark & \cmark & \cmark & \cmark & \cmark & \cmark \\
        Embodied ADP (kg Sb-eq) & --     & \cmark & --     & \cmark & --     & \cmark \\
        \bottomrule
    \end{tabular}
    \caption{Lifecycle phases covered by each reported indicator (\cmark{} included, -- excluded; op.\ = operational, emb.\ = embodied).}
    \label{tab:supp-impact-boundary}
\end{table}

\subsection{Estimating the initial annual volume of generated videos}\label{supp:volumetry}%

The initial volume $V(t_0)$ is the model's dominant input, and no provider
publishes an authoritative generation count.
The evolutions of $V(t)$ are scenario assumptions
(Supplementary Methods~\ref{supp:uq}).
We therefore triangulate two independent sources: our own scrapes of two
platforms, and the figures that providers disclose.
We combine those figures into the triangular distribution described at the end of this subsection.

\textbf{Sora scrape.}
Sora (sora.chatgpt.com) maintained a public feed that listed every public
video with its resolution, duration and prompt; OpenAI discontinued the service
in April 2026. 
We crawled this feed over 54 days between February and April 2026 and recorded
about 3.3 million videos, an average of 61{,}000 per day, or 22M per year.
The crawl covers public videos only, so the figure is a lower bound on
Sora's output.
The same corpus provides the resolution and duration mixes that weight
per-video latencies (Supplementary Methods~\ref{supp:headline}) and the power-law
file-size model used for storage (Supplementary Methods~\ref{supp:variables}); the
clips are predominantly short and low-resolution, with 76\% lasting five
seconds and 80\% generated at 480p.

\textbf{HeyGen scrape.}
HeyGen displays a cumulative counter of generated videos on its homepage.
We sampled the counter at regular intervals of about five minutes between March and June 2026, and
computed the mean daily generation rate from consecutive counter readings: 304{,}000
videos, or about 111M per year. %
Unlike the Sora crawl, the counter includes private generations, but it
measures a single platform.
It tracks HeyGen's text-to-video service only; the company's
avatar-based and template-driven talking-head products are not included
in the count.
However, that service is not tied to a single engine: the platform's editor routes
prompts both to HeyGen's in-house models and to third-party closed models
that HeyGen reaches through their vendors' APIs, including Sora, Google
Veo and Kling~\cite{heygenAiVideoGenerator2026}.
A share of the counted output is therefore generated outside HeyGen's own
infrastructure, so the 111M figure measures the service's activity rather
than the load on HeyGen's fleet; it may also overlap with the Kling
disclosure below if Kuaishou's count includes API usage, an overlap
bounded at 111M videos per year, about 2\% of the observable total.

\textbf{Counting units.}
The provider disclosures below do not share a common unit, so we normalise them to
single-output equivalents: one video is one finished output, and
multi-variation renders are divided by the number of variations.
The division counts prompt-equivalents: each rendered
variation consumes compute, so per-prompt-equivalent energy implicitly
amortises discarded variations across the market aggregate.

\textbf{Other providers and public disclosures.}
To our knowledge, no other provider publishes a continuing generation count,
but several disclose cumulative figures or recent increments from which
annual volumes can be extrapolated.
\autoref{tab:supp-volume-disclosures} lists the disclosures we use.

Kuaishou reported more than 600M videos for its Kling model over its first
19 months, about 30M per month at the end of 2025~\cite{kuaishouKlingARR240M2026}.
Google reported 275M Veo videos in Flow over five months, about 55M per
month~\cite{gallegosIntroducingVeo312025}.
MiniMax stated that its video models had generated more than 600M videos
cumulatively by the end of 2025, about 17 months after the August 2024
launch of its video generation service, a lifetime average of about 35M
per month~\cite{minimaxAnnualResults2026}.
Shengshu reported more than 400M Vidu videos since the April 2024 launch,
about 22M per month~\cite{shengshuViduQ2Launch2025}.
Higgsfield, a platform that also resells third-party video models, reported
more than 800M pieces of AI media generated on its platform over
January--June 2026~\cite{higgsfieldStateOfAIVideo2026}; its own cumulative
split of 300M videos out of 850M generations puts the video share at about
35\%, which we adopt as about 560M videos per
year~\cite{higgsfieldAbout2026}.
Because the platform resells models that the provider rows also count, this
figure may overlap with the rows above, and we treat it as an upper bound on
the volume Higgsfield adds on top of them.
Grok Imagine (xAI) reported 1.245B rendered clips over the 30 days before
2 February 2026~\cite{xaiGrokImagineVolume2026}, and close to 50M videos per
day at its public all-hands meeting later that month~\cite{xaiAllHands2026}.
These counts cover rendered clips, about four variations per prompt, so we
divide them by four: at single-output equivalents, the disclosed 30-day rate
implies about 3.8B videos per year and the all-hands rate about 4.6B.

\begin{table*}[t]
    \centering
    \resizebox{\linewidth}{!}{%
        \begin{tabular}{@{}lll@{}}
            \toprule
            Provider & Disclosure & Annualised \\
            \midrule
            Kling (Kuaishou) & 600M cumulative over 19 months (Dec 2025); $\approx$30M/mo end-2025 & $\approx$360M/yr \\
            Veo in Flow (Google) & 275M over 5 months (2025) & $\approx$660M/yr \\
            MiniMax (Hailuo) & $>$600M cumulative by Dec 2025 ($\approx$17 months after the Aug 2024 launch) & $\approx$430M/yr \\
            Vidu (Shengshu) & $>$400M cumulative (Oct 2025, since Apr 2024) & $\approx$267M/yr \\
            Higgsfield & $>$800M AI-media pieces (Jan--Jun 2026); video share $\approx$35\% (300M/850M cumulative) & $\approx$560M/yr \\
            Grok Imagine (xAI) & 1.245B rendered clips over 30 days (Feb 2026); $\approx$4 variations per prompt & $\approx$3.8B/yr at single-output equivalents%
             \\
            \addlinespace
            PixVerse & 0.8B $\rightarrow$ 2.1B cumulative, self-reported (Sep 2025 $\rightarrow$ Mar 2026); definition unstated & $\approx$0.8B/yr %
            \\
            \bottomrule
        \end{tabular}}
    \caption{Provider volume disclosures used to build the observable total
    and to justify the high end of the volume prior. Cumulative totals
    are annualised from the most recent disclosed increment, except
    MiniMax, which is annualised from its lifetime total.
    All rows except PixVerse enter the 6.2B observable
    total; PixVerse only justifies its upper tail.}
    \label{tab:supp-volume-disclosures}
\end{table*}

Several providers disclose no usable volume figure.
ByteDance discloses no generation counts for Seedance, and OpenAI never
disclosed Sora volumes before discontinuing the service in April
2026~\cite{openaiSoraDiscontinuation2026}.
To our knowledge, no Chinese provider discloses a monthly rate, only
cumulative totals, so the Kling and Vidu annualisations assume that each
provider's last observed rate held over the base year.
Because these platforms launched between 2024 and 2025, no multi-year
historical trend exists for them: the disclosed rates describe current
annualised output, not a trajectory that can be extrapolated
onwards.

\textbf{Measured floor and prior.}
Our two scrapes put the directly measured output at about 133M videos per
year. Adding the observable disclosures of
\autoref{tab:supp-volume-disclosures} gives
an observable total of about 6.2 billion single-output videos per year
Under our normalisation, the disclosed Grok Imagine rate alone exceeds the
sum of all other observable providers, so a single service accounts for
most of the observable total.
We anchor the prior on this total:
$V(2026)\sim\text{triangular}(6.2\text{B}, 6.2\text{B}, 9.5\text{B})$,
a one-sided distribution. The low bound is
a deliberate undercount: the Sora figure counts public videos only,
several large services entered the market during 2026, and the disclosed
rates are 2025 snapshots.

We justify the 9.5B high end item by item rather than by hand.
Four items add about 3.3B on top of the 6.2B mode.
First, the Grok Imagine upside: the all-hands rate reported above would
add about 0.8B single-output videos per year over the disclosed 30-day
rate; because the platform reached this scale during the base year and no
post-February 2026 figure exists, we keep this increment out of the mode.
Second, PixVerse: its counting definition is unstated, so we count its
self-reported volume at a 70\% discount, which adds about 0.8B videos per
year; the discount is documented in our replication package.
Third, ByteDance: no primary disclosure exists for Seedance, whose video
generation is embedded in Doubao, Jimeng, and CapCut, so we count it at
the midpoint of a Kling-to-Grok parity range, about 1.0B videos per year;
this is a stated analogy, not a disclosure.
Fourth, Google's non-Flow surfaces: we count video generation through the
Gemini app, free YouTube Shorts Remix access, and the Veo API at parity
with Veo in Flow's disclosed rate, which adds about 0.7B videos per year.
The itemised sum is about 9.5B.
Beyond this total we cannot quantify several possible additions: Meta
Vibes, Midjourney's unlimited-tier generations, and the further long tail.
The Kling rate is also a 2025 figure; Kuaishou's reported revenue growth for
Kling (annualised revenue run rate up from USD100M in March 2025 to USD240M
in December 2025) suggests the underlying volume grew further, but no
updated count exists~\cite{kuaishouKlingARR240M2026}.

\textbf{Average-to-peak utilisation.}
The utilisation $\tauInf$ of the inference fleet is the fraction of provisioned
GPU time spent actively generating.
We estimate it from the same scrapes as the ratio of average to peak
generation rate, where the peak is the mean of the five busiest hourly bins:
Sora yields 62.1\% and HeyGen 45.8\%.
The two scrapes give point estimates for two services only, so we do not fix $\tauInf$ at either value: the 2026 value is drawn from $\tauInf\sim\text{triangular}(0.5, 0.63, 0.75)$, a prior whose mode sits on Sora's measured 62.1\%, the higher of the two. The lower and upper bounds are set by hand to span preferably higher-utilisation services; in particular, the low bound does not cover HeyGen's 45.8\%.

\subsection{Impact estimation model structure}\label{supp:model_structure}
The model estimates the yearly environmental impact of a text-to-video generation service through a pipeline of calculations involving 32 input variables and 11 output variables, with results for year $t$ feeding year $t+1$ through, for example, the GPU fleet composition.
Its dominant path runs from the annual video volume $V(t)$ through per-video latencies, annual active GPU-hours, and inference, training, and storage energy, to facility electricity and the three reported impacts.
This section gives the equations, coefficients, and per-step definitions; \autoref{tab:supp-notation} summarises the notation.

Three assumptions bound what the model can represent.
First, generation models are treated as black boxes characterised only by
video quality and cost, so GPU memory requirements are not modelled.
Second, a GPU handles exactly one inference at a time and one inference occupies
exactly one GPU: we found no published evidence of batched video generation at
production scale, and splitting a model across GPUs improves latency but not
throughput~\cite{jeghamLightsCameraCarbon2026}.
Third, only inference GPUs form an explicit fleet with a purchase and
retirement schedule; training hardware is accounted for by amortising its
GPU-hours, without modelling a purchase cadence or bursty peak concurrency.
Training is likewise a sequence of discrete runs and not a continuous workload:
video-generation models are released as versions, each from a bounded
training campaign, and the work between releases is covered by the
development-overhead multiplier of \autoref{supp:headline}.

\begin{table*}[t]
    \centering
    \caption{Core notation used in the Supplementary Methods.}
    \label{tab:supp-notation}
    \rowcolors{2}{}{gray!12}%
    \begin{tabular}{@{}lll@{}}
        \toprule
        Symbol & Meaning & Unit \\
        \midrule
        $t$ & Simulation year & year \\
        $V(t)$ & Annual generated video volume & video/year \\
        $\latency(\videoRes,u,m)$ & Per-video latency for resolution $\videoRes$ and duration $\videoDur$ on GPU model $m$ & s/video \\
        $\Theta_{\gpuModel}$, $S$ & Peak BF16 dense throughput of GPU model $m$; denoising steps & TFLOP/s, -- \\
        $\alpha$, $\beta$, $\gamma$ & Fitted latency coefficients (quadratic, linear, constant) & -- \\
        $\resMix$, $\durMix$ & Resolution and duration shares & -- \\
        $\gpuMix$ & GPU fleet share for model $m_i$ & -- \\
        $I$, $J$, $K$ & Numbers of GPU models, video durations, and resolutions & -- \\
        $\eta_{\text{opt}}$ & Inference-optimization factor & -- \\
        $\tauInf$, $\tauTrain$ & Inference and training GPU utilisation rate & -- \\
        $H_{\text{inf}}$, $H_{\text{train}}$ & Active inference and training GPU-hours & GPU h \\
        $\hat{H}_\infSuffix(t,m)$ & Active inference GPU-hours on GPU model $m$ & GPU h \\
        $\nbGpuInf$, $\nbGpuTrain$ & Required inference and training GPUs & GPU \\
        $\nrj_\infSuffix$, $\nrj_\trainSuffix$, $\nrj_\storSuffix$ & Annual inference, training, and storage energy & MWh \\
        $\nrjInfAct$ & Active inference energy, idle draw excluded & MWh \\
        $e_\infSuffix(t)$ & Average per-video inference energy & Wh/video \\
        $\nbTrain$, $\nrjPerTrain$, $\devOverhead$ & Training runs per year, energy per training run, development overhead & --, MWh, -- \\
        $\trainModel$, $P(m)$, $P_{\text{idle}}(m)$ & Training GPU model; nameplate and static GPU power & --, W, W \\
        $\powerRateTrain$ & Training power fraction of the GPU nameplate power & -- \\
        $P_{\text{avg}}(t)$ & Fleet-weighted average inference GPU power & W \\
        $N_i(t)$ & Provisioned inference fleet of GPU model $m_i$ & GPU \\
        $d_{\text{GPU}}$ & GPU lifespan & years \\
        $\text{gwp}_{\text{emb}}$, $\text{adp}_{\text{emb}}$ & Per-GPU embodied GHG and ADP coefficients & kg CO$_2$eq, kg Sb-eq \\
        $\eta_{\text{srv}}$, $\eta_{\text{it}}$, PUE & Server and IT-equipment overheads; power usage effectiveness & -- \\
        $\text{WUE}$ & Water Usage Effectiveness & L/kWh \\
        $S_{\storSuffix}(t)$, $e_{\text{stor}}$ & Stored video data; storage electricity intensity & TiB, MWh/GiB \\
        $\rho$ & Storage replication factor & -- \\
        fps & Frame rate of generated videos & frames/s \\
        $E_{\text{IT}}(t)$, $E(t)$ & IT-equipment energy; facility electricity & MWh, GWh \\
        $P_{\text{peak}}(t)$ & Peak facility power & MW \\
        $G(t)$, $W(t)$, $M(t)$ & GHG, operational water, and embodied abiotic depletion potential & t CO$_2$eq, kL, kg Sb-eq \\
        \bottomrule
    \end{tabular}
\end{table*}

\subsection{Headline relations}\label{supp:headline}

Seven groups of relations fix our model's outputs.

\textbf{Per-video inference latency.}
The model's primary physical quantity is the \emph{per-video generation
latency}%
$\latency(\videoRes,u,m)$,
the number of GPU-seconds of model $m$ needed to generate one video of resolution $\videoRes$ and duration $\videoDur$.
Latency follows a compute-bound model for DiT text-to-video diffusion taken
from Delavande \etal~\cite{delavandeVideoKilledEnergy2025}: per-video latency is
quadratic in the DiT token length, and therefore grows with resolution,
duration, and frame rate.
The DiT token length of a video of resolution $\videoRes$ and duration $u$ is
$\ell(\videoRes,u) = \bigl(1 + u\,\text{fps}/4\bigr)\,(h/16)\,(w/16)$, where $h$ and $w$ are the pixel height and
width of $\videoRes$, each rounded to the nearest multiple of 16 pixels (the width follows from the height at a 16:9
aspect ratio), and fps is the frame rate. The latency of that video on GPU model $m$ is
\begin{equation}
    \latency(\videoRes, u, m) = \bigl(\alpha\,S\,\ell(\videoRes,u)^2 + \beta\,S\,\ell(\videoRes,u) + \gamma\bigr)
    \cdot \frac{\Theta_{\text{H100}}}{\Theta_m} \cdot \eta_{\text{opt}},
    \label{eq:supp-latency}
\end{equation}%
where $\alpha$, $\beta$, $\gamma$ are the fitted coefficients, $S$ is the
number of denoising steps, fixed at 50 in all scenarios, $\Theta_{\gpuModel}$ is
the peak throughput of GPU model $\gpuModel$, $\Theta_{\text{H100}}$ its value for the H100 (the configuration the
coefficients were fitted on), and $\eta_{\text{opt}}$ an inference-optimization factor defined below.
The quadratic term is self-attention, the linear term the MLP and
cross-attention layers, and the constant a fixed per-generation overhead.
The linear and constant coefficients are fitted on WAN2.1-T2V-1.3B measurements
taken on H100 by Delavande \etal~\cite{delavandeVideoKilledEnergy2025} and shared across generation models, while the quadratic
coefficient $\alpha$ is refitted per generation model; all our scenarios use
WAN2.1-T2V-1.3B. The fitted latencies are transferred to other GPU models by
scaling with the ratio of peak BF16 dense throughputs: a B200 at 2250~TFLOP/s is
modelled as $2250/989\approx2.27$ times faster than an H100.
This transfer is sound
only in a compute-bound regime and would break down if
generation were bottlenecked by memory bandwidth.

The fitted latency describes an unoptimized research configuration, not a
deployed endpoint.
A single \emph{inference-optimization factor}
$\eta_{\text{opt}}>0$ closes that gap, absorbing step and
classifier-free-guidance distillation, kernel fusion, caching, sparse attention
and quantization into one multiplier on latency, and, since energy consumption during video generation is
latency times power, on per-video energy as well. %

We call the prediction of \eqref{eq:supp-latency} at $\eta_{\text{opt}}=1$ the
\emph{unoptimized baseline}: the WAN2.1-fitted coefficients applied as measured,
with no production inference optimization.
Factors above 1 are possible: they describe models whose serving configuration is less optimized than this baseline.
We calibrate $\eta_{\text{opt}}$ against one of the per-clip energy figures reported by
Jegham \etal~\cite{jeghamLightsCameraCarbon2026}, who estimate text-to-video
inference energy from architectural first principles and observable generation
parameters.
On our reference workload (one 8s, 720p clip at 24~fps on a single H100), the
unoptimized baseline
 predicts 366~Wh, and matching Seedance~1.5's 102.1~Wh gives
$\eta_{\text{opt}}=0.279$ for the base year. %
\autoref{tab:supp-optimization-factors} reports the factor implied by every model in
that survey.
Both $\eta_{\text{opt}}$ and the resolution and duration mix are per-year inputs, so $\latency$ carries
an implicit dependence on $t$ that we leave out of the notation; the yearly trajectory of $\eta_{\text{opt}}$ is
quantified in Supplementary Methods~\ref{supp:uq}.

\textbf{Inference GPU-hours and GPU count.}
Let $\hat{H}_\infSuffix(t,m)$ be the annual GPU-hours required to generate the $V(t)$ yearly videos on GPU model $\gpuModel$ alone.
We consider $J$ video durations $\videoDur_j$ ($j \in [1..J]$) with respective shares $\durMix(t)$, and $K$ resolutions $\videoRes_k$ with respective shares $\resMix(t)$; each set of shares sums to one, $\sum_{j=1}^J \durMix(t) = \sum_{k=1}^K \resMix(t) = 1$, and, for the sake of simplicity, the resolutions and durations are assumed independent.
$\hat{H}_\infSuffix(t,m)$ is computed from the per-video latency of \eqref{eq:supp-latency}, weighted by these mixes.
Putting this together, we obtain the annual GPU-hours as:
\begin{equation}
    \hat{H}_\infSuffix(t,m) = \frac{V(t)}{3600}
    \sum_{k=1}^{K} \sum_{j=1}^{J}
    \resMix(t)\, \durMix(t)\, \latency(\videoRes_k, u_j, m),
    \label{eq:supp-gpu-hours-per-model}
\end{equation}

where the double sum represents the average per-video latency under the current year's resolution and duration mixes for a given GPU model. The factor $1/3600$ converts GPU-seconds to GPU-hours.
Taking into account the $I$ GPU models $\gpuModel_i$ with respective share $\gpuMix$ (with $\sum_{i=1}^I \gpuMix(t) = 1$), we obtain the annual active GPU-hours for inference $H_{\text{inf}}(t)$ as:
\begin{equation}
    H_{\text{inf}}(t) = \sum_{i=1}^{I} \gpuMix(t)\, \hat{H}_\infSuffix(t, m_i).
    \label{eq:supp-gpu-hours}
\end{equation}

The size of the inference fleet follows from the utilisation rate $\tauInf$, the fraction of provisioned GPU time spent actively generating (Supplementary Methods~\ref{supp:volumetry}); the remaining fraction $1-\tauInf$ is idle time, during which GPUs have no job.
A lower $\tauInf$ inflates the fleet required to serve the same video volume.
The required inference fleet is the ratio between the required annual GPU-time and the active GPU-time offered by one GPU over a year:
\begin{equation}
    \nbGpuInf(t) = \left\lceil
    \frac{H_{\text{inf}}(t)}{8760 \times \tauInf} \right\rceil.
    \label{eq:supp-required-gpu}
\end{equation}
Here $8760$ is the number of hours in a year and the ceiling reflects that a
fractional GPU cannot be provisioned.
Later years scale the 2026 value of $\tauInf$ by each scenario's annual utilisation gain, compounded year on year; the gains are quantified in Supplementary Methods~\ref{supp:uq}.

\textbf{Inference energy consumption.}
In terms of energy consumption, each GPU model $m$ is characterised by its nameplate power $P(m)$ at active hours, and its static power $P_{\text{idle}}(m)$ at idle hours.
In our simulations, $P(m)$ is the thermal design power (TDP) provided by the manufacturers.
The active part $\nrjInfAct(t)$ (in MWh) is computed from the previous active GPU-hours spread over the mix of GPU models:
\begin{equation}
    \nrjInfAct(t) =
    10^{-6} \sum_{i=1}^{I} \gpuMix(t)\, \hat{H}_\infSuffix(t, m_i) \, P(m_i)
    \label{eq:supp-inference-active}
\end{equation}
In \eqref{eq:supp-inference-active}, the factor $10^{-6}$ converts Wh to MWh.
Adding idle consumption yields the total inference energy consumption of the GPU fleet:
\begin{align}
    \nrj_\infSuffix(t) = & \, \nrjInfAct(t) \nonumber\\[1pt]
    & + \frac{H_\infSuffix(t)}{10^6}\, \frac{1-\tauInf}{\tauInf}
    \sum_{i=1}^{I} \gpuMix(t)\, P_{\text{idle}}(m_i) .
\label{eq:supp-total-inference-energy}
\end{align}

Finally, we also compute the average per-video energy $e_\infSuffix(t) = 10^{6}\,\nrjInfAct(t)/V(t)$, in Wh per video, from the active inference energy alone. It is reported alongside the annual totals and used in the Results.

\textbf{Training GPU count and energy.}
In contrast to the inference fleet, for training we assume a homogeneous fleet of GPUs of model $\trainModel$, running at a fraction $\powerRateTrain$ of the nameplate power $P(\trainModel)$. This fraction, like the utilisation rate $\tauTrain$ below, is an explicit modelling assumption without external source (Supplementary Methods~\ref{supp:variables}).
Idle time is captured as before, through the training utilisation rate $\tauTrain$: the fraction of provisioned training GPU-time spent actively computing, in analogy with $\tauInf$.
We consider $\nbTrain$ training runs per year, each requiring $\nrjPerTrain$ MWh on GPUs of model $\trainModel$.
Training is modelled per run because published evidence reports training campaigns as per-run totals: announced
GPU-hour counts (converted to energy at nameplate power) or training-FLOP estimates. A per-run figure already
integrates a campaign's duration, hardware, and utilisation, so the run energy needs no fleet profile for a workload
that runs in bounded bursts; the per-run energy is converted back into GPU-hours only to size the training fleet
and its idle draw, at the assumed training power fraction $\powerRateTrain$ rather than the campaign's own utilisation.
Research and development overhead is captured through a $\devOverhead$ additional factor: one training run induces an additional cost equivalent to $\devOverhead$ training runs, \ie a total of $(1+\devOverhead)$ runs.

The number of training GPU-hours is computed as the ratio between the estimated total energy demand for the $\nbTrain$ training runs per year, including research and development overhead, and the \textit{useful} energy a single training GPU can deliver in one hour:
\begin{equation}
    H_{\text{train}}(t) =
    \frac{\nbTrain \, \nrjPerTrain(1+\devOverhead) \, 10^6}
    {\powerRateTrain  P(\trainModel)}.
    \label{eq:supp-training-hours}
\end{equation}
The factor $10^6$ converts MWh to Wh.
Taking into account idle time, the training GPU count is obtained as for the inference one, i.e.:
\begin{equation}
    \nbGpuTrain(t) = \left\lceil
    \frac{H_{\text{train}}(t)}{8760 \times \tauTrain}
    \right\rceil.
    \label{eq:supp-training-gpus}
\end{equation}
The \textit{GPUs Required} indicator reported in the results is the sum of $\nbGpuInf(t)$ and $\nbGpuTrain(t)$, though only $\nbGpuInf(t)$ explicitly drives hardware purchases in our model.
The sizing in \eqref{eq:supp-training-gpus} spreads the year's training GPU-hours uniformly across the 8760 hours: $\nbGpuTrain(t)$ is a sizing construct for the reported fleet, not a description of a bursty training campaign.
Annual training energy is the sum of the active training energy and idle time power consumption:
\begin{align}
    \nrj_\trainSuffix(t)=&\, \nbTrain \,\nrjPerTrain(1+\devOverhead) \nonumber\\[1pt]
    &+ \frac{H_{\text{train}}(t)}{10^6}\, \frac{1-\tauTrain}{\tauTrain}\, P_{\text{idle}}(\trainModel) .
    \label{eq:supp-training-energy}
\end{align}

\textbf{Storage}.
The stored volume $S_{\storSuffix}(t)$ (in TiB) is the product of each year's video volume and that year's average file size, accumulated over the rolling retention window and scaled by the replication factor $\rho$ (Supplementary Methods~\ref{supp:variables}). The annual storage energy is then
\begin{equation}
    \nrj_\storSuffix(t) = e_{\text{stor}} \cdot 1024 \cdot S_{\storSuffix}(t),
    \label{eq:supp-storage-energy}
\end{equation}
in MWh, where the factor 1024 converts TiB to GiB and $e_{\text{stor}} = 6.6\times10^{-6}$~MWh/GiB (\ie 6.6~Wh/GiB, converted from the source's 6.132~kWh per decimal TB per year) is the annual electricity intensity of HDD storage systems~\cite{shehabiDataCenterEnergyUsage2024}.

\textbf{Total energy consumption and peak power demand.}
GPU-board and storage energy are raised to facility level through nested
overheads: a server factor, an IT-equipment factor, and the Power Usage Effectiveness (PUE) factor covering the remaining non-IT loads,
following Epoch AI's methodology~\cite{EpochGPUPowerUsage2025}.
Storage energy receives only PUE: its coefficient is already an IT-level figure
that counts the storage servers themselves, so applying the server and
IT-equipment overheads would double-count them.
The IT-equipment energy $E_{\text{IT}}(t)$ and the facility electricity $E(t)$ are then
\begin{equation}
    \begin{split}
        E_{\text{IT}}(t) = {} & \eta_{\text{srv}}\, \eta_{\text{it}}\,
        \bigl( \nrj_\infSuffix(t) + \nrj_\trainSuffix(t) \bigr) + \nrj_\storSuffix(t), \\
        E(t) = {} & \frac{\text{PUE} \cdot E_{\text{IT}}(t)}{10^3},
    \end{split}
    \label{eq:supp-facility-energy}
\end{equation}
in MWh and GWh respectively.
Peak facility power is sized for the worst case in which the whole provisioned
inference fleet and the whole provisioned training fleet draw full nameplate power at once:
\begin{equation}
    \begin{split}
        P_{\text{peak}}(t) = \Bigl[ {} &
        \eta_{\text{srv}} \eta_{\text{it}}
        \frac{\nbGpuInf(t)\, P_{\text{avg}}(t) + \nbGpuTrain(t)\, P(\trainModel)}{10^6} \\[2pt]
        & + \frac{\nrj_\storSuffix(t)}{8760} \Bigr] \cdot \text{PUE},
    \end{split}
    \label{eq:supp-peak-power}
\end{equation}
in MW, where $P_{\text{avg}}(t)=\sum_i \gpuMix(t) P(m_i)$ is the fleet-weighted
nameplate power, $\eta_{\text{srv}}$ and $\eta_{\text{it}}$ are the server and
IT-equipment overheads, and $\nrj_\storSuffix(t)$ is the storage energy, which
runs continuously and therefore peaks at its average.
This is a lower bound on the true peak: $\nbGpuTrain(t)$ spreads the year's
training GPU-hours uniformly across 8760~hours, so a bursty training campaign
concentrating the same hours into a shorter window would need more simultaneous
GPUs.

\textbf{Embodied impacts.}
Embodied impacts are amortised as a \emph{stock}, spread over the provisioned fleet and the GPU lifespan, rather than charged in full in the purchase year.
This attributes each GPU's manufacturing footprint to the years it is
actually in use, avoiding a spike disconnected from the fleet's operation.
For GHG, the inference fleet contributes
\begin{equation}
    G_{\text{gpu}}(t) = \frac{1}{1000} \sum_i
    \frac{N_i(t)\, \text{gwp}_{\text{emb}}(m_i)}{d_{\text{GPU}}},
    \label{eq:supp-embodied}
\end{equation}
in t~CO\textsubscript{2}eq, where $\text{gwp}_{\text{emb}}(m_i)$ is the per-GPU embodied GHG coefficient in
kg CO$_2$eq, $N_i(t)=\gpuMix(t)\, \nbGpuInf(t)$ is the provisioned fleet of model $m_i$, $d_{\text{GPU}}$ the GPU
lifespan in years, and the factor $1/1000$ converts kg to tonnes.
Abiotic depletion potential follows the identical structure with
$\text{adp}_{\text{emb}}(m_i)$, the per-GPU ADP coefficient, in place of $\text{gwp}_{\text{emb}}(m_i)$.
Training hardware is charged the same way but amortised over GPU-hours rather
than GPU-years, as it is not provisioned through the fleet's purchase and
retirement path; because the training GPU-hours inherit the development-overhead
multiplier, so does the training embodied impact.
The model aggregates these terms into the three reported impacts:
GHG emissions $G(t)$ sum the operational emissions of facility electricity and the embodied terms of the GPU fleet, storage hardware, and training hardware.
Embodied ADP $M(t)$ sums the same three embodied terms, with per-GPU ADP coefficients in place of the GHG ones; it carries no operational component.
Operational water $W(t)$ multiplies the IT-equipment energy $E_{\text{IT}}(t)$ of \eqref{eq:supp-facility-energy} by
the Water Usage Effectiveness (WUE), the facility water input per unit of IT energy. The inventory counts direct
potable and non-potable water only; indirect water, and the water embodied in hardware manufacturing, stay outside
it.%

\subsection{Hardware, parameters, and data sources}\label{supp:variables}

We model two accelerators, H100 and B200, deployed in 8-GPU DGX-class nodes.
These two bracket the hardware actually available to a 2026 service: the
intermediate B100 was effectively cancelled over a design flaw and most cloud
providers moved straight from Hopper to the B200~\cite{garreffaAnalystNvidiaEffectivelyCanceled2024}.
Embodied figures sum the bare accelerator board, taken from the Hubblo GPU
dataset~\cite{leesPerassoGpuLcaAdeme2026}, and a per-GPU (1/8)
share of its host server, obtained from BoaviztAPI~\cite{simonBoaviztapiBottomUp2024};
\autoref{tab:supp-gpu-hardware-specs} gives the resulting per-GPU values.
These embodied figures should be read as carrying at least a factor-of-two
uncertainty: BoaviztAPI's own GPU-component model returns roughly twice the
Hubblo board figure for both accelerators, and we have not resolved the
divergence.

\textbf{What the ADP figure contains.} The two datasets publish different ADP
variants: the Hubblo board values are the abiotic depletion potential for
mineral and metal resources (ADPe), while the BoaviztAPI host-server and
hard-drive values are the combined ADP, which BoaviztAPI documents as covering
both a material and a fossil-resource component.
For the H100, the host term accounts for 0.04875~kg Sb-eq of the total 0.0577~kg
Sb-eq per GPU, and about a tenth of that host value is its mineral component, so roughly
three quarters of the per-GPU figure reflects fossil-resource depletion rather
than minerals.

We keep the combined ADP rather than substitute a mineral-only figure because
a configuration-sensitive one exists for the board but not for the host server
or the hard drive: BoaviztAPI returns the same value for every server
configuration and none at all for storage hardware.
The composition matters little at service level: storage stays below 0.1\% of
the total in every reported year, and the inference fleet supplies roughly
60\% or more of it, so the indicator is dominated by the embodied ADP of the
inference fleet.

\begin{table}[t]
    \centering
    \caption{Per-GPU hardware specifications (board plus a 1/8 share of the
        DGX host). Peak throughput is BF16 dense and drives the cross-GPU
        latency transfer. Idle power feeds the inference and training
        idle-energy terms.}
    \label{tab:supp-gpu-hardware-specs}
    \resizebox{\columnwidth}{!}{%
        \begin{tabular}{@{}lccccc@{}}
            \toprule
            Model & TDP (W) & Idle (W) & Peak (TFLOP/s) & GHG (kg CO$_2$eq) & Embodied ADP (kg Sb-eq) \\
            \midrule
            H100  & 700     & 59       & 989            & 885.5             & 0.0577              \\
            B200  & 1000    & 140      & 2250           & 1321.5            & 0.05896             \\
            \bottomrule
        \end{tabular}}
\end{table}

\autoref{tab:supp-optimization-factors} lists the per-clip energies that Jegham
\etal report for each model of their survey and the optimization factor each
one implies on our reference workload.

\begin{table}[t]
    \centering
    \caption{Inference-optimization factor $\eta_{\text{opt}}$ implied by each
        model surveyed by Jegham \etal~\cite{jeghamLightsCameraCarbon2026}, on a
        reference workload of one 8\,s 720p clip at 24~fps on a single H100. Our
        unoptimized baseline predicts 366~Wh for that clip. The base scenario
        adopts the Seedance~1.5 value.}
    \label{tab:supp-optimization-factors}
    \begin{tabular}{@{}lcc@{}}
        \toprule
        Model & Per-clip energy (Wh) & $\eta_{\text{opt}}$ \\
        \midrule
        Veo 3            & 30.8  & 0.084 \\
        Veo 3.1          & 39.5  & 0.108 \\
        Seedance 1       & 80.0  & 0.219 \\
        \textbf{Seedance 1.5} & \textbf{102.1} & \textbf{0.279} \\
        Wan 2.1-14B      & 382.7 & 1.045 \\
        HunyuanVideo     & 411.4 & 1.124 \\
        HunyuanVideo-1.5 & 478.5 & 1.307 \\
        \bottomrule
    \end{tabular}
\end{table}

Idle power is rarely published by manufacturers, so both figures come from
independent measurements: 59~W for the H100 from an NVML-instrumented inference
benchmark~\cite{argerichWattCounts2026}, and 140~W for the
B200 from a reported \tech{nvidia-smi} reading on an 8-GPU
node~\cite{lightlyB200Benchmark2025}.
GPU lifespan is modelled as $d_{\text{GPU}}\sim\text{uniform}(3,7)$ years,
spanning the tension between hyperscaler depreciation schedules (4--6 years)
and the shorter service lives implied by observed failure rates and
obsolescence.

Annual video volume follows the prior described in
Supplementary Methods~\ref{supp:volumetry}.
Training energy per run is treated as uncertain across the range of published
2024-and-later text-to-video models,
$e_{\text{train}}\sim\text{triangular}(130, 465, 1800)$ MWh.
The mode is Seaweed-7B's reported 665{,}000 H100 GPU-hours ($\approx465$ MWh at
full TDP), one of the only primary-sourced GPU-hour figures among commercial models and
the median of the models we surveyed; the low bound is an efficient 14B model of
the Wan class and the high bound a 30B frontier model, with the spread driven
almost entirely by parameter count~\cite{EpochAIDataOnAIModels2026}.
Published GPU-hour counts are read as energy at full TDP and converted back to
GPU-hours at the 80\% training power assumed below, so the modelled GPU-hours
exceed the published count by a factor $1/0.8$. Energy is preserved; the
provisioned training count and its idle draw follow the inflated hours.
The yearly retraining count is $\nbTrain\sim\text{uniform}(20,40)$, the product of
the number of laboratories that train AI video-generation models and one to
two trainings per laboratory per year;
Supplementary Methods~\ref{supp:retraining} derives this prior.
The development
overhead is $k_{\text{dev}}\sim\text{triangular}(0.5,3.0,6.0)$, anchored on the
OLMo~\cite{morrisonHolisticallyEvaluatingEnvironmental2025} and
Moshi~\cite{lopezRauhutEnvironmentalFootprintGenAI2026} footprint studies: the former
reports a development phase costing about half the final run for a mature
pipeline, the latter up to forty times for a radically novel architecture.
Training is assumed to run on H100s at 80\% of TDP, with GPUs active 70\% of
their provisioned time.

Storage impact is driven by a per-video file size, itself a power law fitted on
the same Sora scrape as the resolution and duration mixes, and by a stored-data volume
accumulated over a rolling retention window and scaled by a replication factor.
Storage energy uses Shehabi et al.'s 6.132 kWh/TB/yr operational
intensity~\cite{shehabiDataCenterEnergyUsage2024}, and storage embodied impacts use
BoaviztAPI's per-drive HDD figure amortised over a 24\,TB nearline drive and a
5-year service life; the reference scenario keeps three replicated copies
($\rho=3$) over a retention window drawn from $\text{uniform}(1,3)$ years.

Facility overheads draw on an EU data-centre
report~\cite{euDataCentresEnergyPerformance2025} for PUE and WUE, and on Epoch AI's GPU
power methodology~\cite{EpochGPUPowerUsage2025} for the server
($\eta_{\text{srv}}=1.53$) and IT-equipment ($\eta_{\text{it}}=1.14$)
multipliers; combined with our PUE, these leave GPU boards accounting for
roughly 40\% of facility power.
PUE is drawn from a Weibull distribution fitted to the EU survey (mean 1.49) and
WUE from $\text{triangular}(0.5, 0.704, 0.9)$ L/kWh (mean 0.70), its mode
anchored on the report's hyperscaler average (0.704 L/kWh, Fig.~12).
Carbon intensity is drawn from a distribution fitted to European grid
data~\cite{eeaGreenhouseGasIntensityElectricity} with a mean of 192~g
CO\textsubscript{2}eq/kWh and a median of 154~g, the gap reflecting a strong
right skew across countries.
Both the PUE and the carbon-intensity distributions are EU-wide and unweighted
by generation location, which is a conservative choice for PUE and an open
question for carbon.

Several inputs carry no external source and should be read as explicit
modelling assumptions rather than fitted quantities: the GPU fleet composition
and its evolution, the training usage and active-time factors, the replication
factor and the retention window.
The retraining count builds on a documented laboratory count, but its retained
value stays an author-set floor
(Supplementary Methods~\ref{supp:retraining}).
The warmup fraction affects only the reported purchase schedule, not any
environmental output, since embodied impacts amortise over the provisioned fleet
rather than the purchase flow.

\subsection{Deriving the yearly retraining count}\label{supp:retraining}

The retraining count combines a count of developing laboratories with a
per-laboratory training rate.
We counted the laboratories that trained and released a video-generation model
for commercial use between 2024 and late 2026; this strict count finds 27
laboratories, and stays between 24 and 34 under other boundary choices (for
example, counting only products still active in late 2026, or also counting
research previews and post-training efforts).
The model retains $N=20$ laboratories.
We set this floor by hand, below the strict count of 27, as a conservative
choice: it absorbs the retirement of training programmes (two
major products were discontinued in 2026), releases limited to early access,
and laboratories whose continued activity we could not verify.
The full laboratory list, with one source per entry, is in our replication
package.
Each retained laboratory trains one to two video-generation models per year,
drawn as $\text{uniform}(1,2)$, and the count is held constant across the
2026--2030 horizon, so the retraining count follows
$\nbTrain\sim\text{uniform}(20,40)$.

\subsection{Scenarios}\label{supp:uq}

All scenarios share a 2026 baseline and a common technology trajectory in which
utilisation improves and inference optimizations compound year over year, and
they differ in how demand and video quality evolve to 2030.
Utilisation rises year on year by $+10\%$ in 2027, $+5\%$ in 2028, $+2\%$ in 2029, and $+0.1\%$ in 2030, and the
inference-optimization factor falls by $20\%$, $10\%$, $5\%$, and $5\%$ over the same years, each change applied
multiplicatively to the previous year's value.

We report three scenarios.
The \emph{Reference scenario} lets demand, resolution and duration all grow, with
the fleet renewing towards B200s.
\emph{Usage stagnation} freezes volume and video quality at their 2026 values,
so that only the technology evolves; it isolates what efficiency gains alone can
achieve.
\emph{Mass adoption} keeps the technology, quality and training trajectories of
the \emph{Reference scenario} and accelerates only the volume path, to test how
the 2030 impacts respond to demand above the \emph{Reference scenario}; because
the per-video quantities are identical in both scenarios, every impact
indicator scales with volume.
The yearly retraining count keeps the prior derived in
Supplementary Methods~\ref{supp:retraining} in all three scenarios, so the
scenarios vary demand rather than the intensity of the training race;
Supplementary Results~\ref{supp:training-constancy} reports a diagnostic
sensitivity run in which this assumption is relaxed.

The volume trajectory of the \emph{Reference scenario} is deliberately modest.
Its $12.8\times$ growth over four years is an average yearly multiple of about
$1.9$: volume doubles in each of 2027, 2028 and 2029, then grows by $+60\%$ in
2030.
Growth at Google ran faster: the company reported that the tokens processed
monthly across its AI products, a broader unit of activity than video
generations, rose from 9.7 trillion in May 2024 to more than
3.2 quadrillion in May 2026~\cite{pichaiGoogleIO2026}.
Against this curve, the expansion of the \emph{Reference scenario} sits below the
growth rates reported by large operators, and we read it as a conservative
baseline rather than a central estimate.
Usage can also contract: OpenAI discontinued Sora in
2026~\cite{openaiSoraDiscontinuation2026}.

The accelerating rates of \emph{Mass adoption} (yearly multiples of 2.0 in
2027, 2.5 in 2028, 4.0 in 2029 and 3.0 in 2030) remain at or below the
expansion reported for text- and image-generation services over 2024--2026:
the Google token growth cited above breaks down into a fifty-fold increase
between 2024 and 2025 and a seven-fold increase between 2025 and 2026,
while ChatGPT reached about 10\% of the
world's adult population within three years of its
launch~\cite{chatterjiHowPeopleUseChatGPT2025}.
Whether even these lower rates transfer to video is an open question, so we
read the scenario as an illustrative sensitivity case rather than a forecast.

\subsection{Uncertainty analysis}
We quantify output uncertainty with a Monte Carlo procedure over scenario
inputs: most numeric fields can be annotated with a distribution (normal,
lognormal, uniform, triangular, or a generic distribution with optional bounds)
in place of a fixed value, and each run draws one sample of every annotated
field before executing the pipeline; the per-element shares of the quality and
fleet mixes are the exception and stay fixed.
All reported results use 10{,}000 Monte Carlo samples with a fixed random seed
(42), for the three scenarios and for the one-at-a-time sensitivity runs, so every
table, tornado panel, and distribution figure in this paper comes from the same
seeded run.
We complement this with a one-at-a-time sensitivity analysis, sampling a single
uncertain field per pass while holding the rest at their median.

A final diagnostic run, not one of the three scenarios, isolates the role of
video quality: it follows the \emph{Reference scenario}'s demand and quality
paths but holds the efficiency-related inputs at their 2026 values, with the
GPU fleet kept at 100\% H100 and the inference-optimization factor and GPU
utilisation frozen. Quality growth alone would raise energy per video from
24.4~Wh in 2026 to $20.9\times$ its 2026 level by 2030, against the $8.8\times$
observed under the full technology trajectory.

\section{Threats to Validity}\label{supp:threats}

This section lists the methodological threats that remain on top of the
deliberate scope decisions recorded in the Limitations subsection of the
Methods section. The items here are uncontrolled issues, not scope
choices, and each is stated with its mitigation and its residual risk.
Because the study is a simulation-based estimate rather than a controlled
experiment, internal validity is read as the validity of the model
mechanics: given the inputs, are the equations and the transfer steps the
correct implementation of the mechanism they claim?

\subsection{Construct validity}

\textbf{The 2026 volume prior.}
The dominant input $V(2026)$ is drawn from a one-sided triangular distribution of
6.2--9.5B videos whose low end and mode sit on the observable total of
measured and disclosed rates, above our measured floor of roughly 133M videos per
year.
The disclosed rates are point-in-time, partial, and self-reported; the two
scrapes behind the floor are only a floor.
\textit{Mitigation:} the prior is deliberately wide, anchored at both
ends, and the floor comes from two independent measurements.
\textit{Residual risk:} high, on the absolute 2026 level of every output;
lower on scenario comparisons, which share the same 2026 baseline.

\textbf{The per-video energy anchors.}
The 24.4~Wh GPU-active energy per video is derived, not measured: a
quadratic latency model fitted on one model (WAN2.1-T2V-1.3B), multiplied
by a single optimization factor that no anchor alone can identify, and
calibrated on one clip (Seedance~1.5, 102.1~Wh).
\textit{Mitigation:} the implementation reproduces the source benchmark to
within 2\%~\cite{delavandeVideoKilledEnergy2025}, the model's unoptimized-baseline
366~Wh prediction for the reference clip falls inside the 30.8--478.5~Wh range
of Jegham \etal~\cite{jeghamLightsCameraCarbon2026}, and the market-cost
validation matches inside the fitted domain.
\textit{Residual risk:} every model in that survey implies a different
optimization factor (0.084--1.307); the inference energy inherits this
spread.

\textbf{One platform's resolution and duration mix.}
The resolution and duration mixes (76\% last five seconds, 80\% at 480p,
treated as independent marginals) and the storage file-size model come
from one platform's public feed (Sora, 3.3M videos).
\textit{Mitigation:} the corpus is large, but the mix is a fixed input
rather than an uncertainty, so its error does not appear in the reported
ranges.
\textit{Residual risk:} if the global mix skews to longer or
higher-resolution clips, per-video energy and storage are both understated;
the risk runs in both directions.

\textbf{European conversion coefficients.}
PUE, WUE, and grid carbon intensity are drawn from European distributions
and applied without weighting by where generations actually run, so every
non-electricity indicator inherits the location assumption.
\textit{Mitigation:} the PUE distribution is fitted to a European
data-centre survey (mean 1.49) and is conservative for PUE.
\textit{Residual risk:} carbon intensity is the largest single driver of
the GHG spread: operational electricity spans a factor of 2.4 between the 5th and 95th
percentiles of the Monte Carlo samples, GHG 17.1.

\subsection{Internal validity}

\textbf{The 2030 extrapolation stack.}
The \emph{Reference scenario} and \emph{Mass adoption} 2030 figures combine three unverified steps:
the latency coefficients are evaluated outside the fitted token-length
range, where the dominant quadratic term amplifies the error; a constant
optimization factor is applied where measured optimizations strengthen
with token length and therefore under-correct at high tiers; and the
H100-to-B200 transfer assumes an empirical efficiency that was measured
only on H100.
\textit{Mitigation:} the 2026 baseline stays inside the validated regime,
and the market-cost validation shows the model diverges upward at higher
tiers.
\textit{Residual risk:} the 2030 impact levels under the \emph{Reference
scenario} and \emph{Mass adoption} are the least supported numbers in the paper, and on the
per-video-energy side it is likely an overestimate; the lower-bound
framing is solid for 2026, but the 2030 level carries this upward model
error.

\subsection{External validity}

\textbf{A snapshot of a fast-moving field.}
The 2026 inputs (fleet, prices, calibrations, grid mix) are a snapshot,
and the field moves fast: new models, new accelerators, and grid
decarbonisation shift between publication and replication.
\textit{Mitigation:} all results are reported as distributions from one
seeded run, and the technology trajectory is an explicit scenario input.
\textit{Residual risk:} any number quoted from this paper decays; the 2030
claims additionally rest on a technology trajectory that is partly
assumed, as stated in the Limitations subsection.

\subsection{Conclusion validity}

\textbf{One-at-a-time sensitivity.}
The tornado panels rank each input varied alone while the rest are held at
their medians, so compounding between volume, quality, and the
optimization factor is invisible to the ranking.
\textit{Mitigation:} the per-output Monte Carlo spreads (2.4$\times$ for
electricity, 17.1$\times$ for GHG, between the 5th and 95th percentiles) come
from the full joint sampling and do
not depend on the one-at-a-time design.
\textit{Residual risk:} the per-input ranking itself can be an artefact of
the design; the directional claims (carbon intensity dominates GHG; the
training-related parameters lead the GPUs and water
outputs) are the ones that rest on it.

\textbf{Reporting on skewed distributions.}
The tables report medians with min and max over 10{,}000 samples; the min
and max columns are single-sample extremes, not distributional ranges, and
the factor of 123 between the two 2030 scenarios is a ratio of medians.
\textit{Mitigation:} the growth claim is supported by a non-overlap check:
the 2030 fifth percentile (1{,}681~kt) lies above the 2026 ninety-fifth
percentile (409~kt).
\textit{Residual risk:} median comparisons can hide overlap where
distributions cross; the paper makes no such crossing claim. The
123$\times$ scenario contrast is itself widened by the extrapolation stack
above, since the \emph{Usage stagnation} trajectory stays in the validated
regime.

\end{document}